ROBOMECH Journal

**RESEARCH ARTICLE**

**Open Access**

# Design and implementation of a maxi-sized mobile robot (Karo) for rescue missions

Soheil Habibian[1*][iD], Mehdi Dadvar[1][iD], Behzad Peykari[1], Alireza Hosseini[1], M. Hossein Salehzadeh[1],
Alireza H. M. Hosseini[1] and Farshid Najafi[2]

**Abstract**

Rescue robots are expected to carry out reconnaissance and dexterity operations in unknown environments comprising unstructured obstacles. Although a wide variety of designs and implementations have been presented within the field of rescue robotics, embedding all mobility, dexterity, and reconnaissance capabilities in a single robot remains a challenging problem. This paper explains the design and implementation of Karo, a mobile robot that exhibits a high degree of mobility at the side of maintaining required dexterity and exploration capabilities for urban search and rescue (USAR) missions. We first elicit the system requirements of a standard rescue robot from the frameworks of Rescue Robot League (RRL) of RoboCup and then, propose the conceptual design of Karo by drafting a locomotion and manipulation system. Considering that, this work presents comprehensive design processes along with detail mechanical design of the robot's platform and its 7-DOF manipulator. Further, we present the design and implementation of the command and control system by discussing the robot's power system, sensors, and hardware systems. In conjunction with this, we elucidate the way that Karo's software system and human–robot interface are implemented and employed. Furthermore, we undertake extensive evaluations of Karo's field performance to investigate whether the principal objective of this work has been satisfied. We demonstrate that Karo has effectively accomplished assigned standardized rescue operations by evaluating all aspects of its capabilities in both RRL's test suites and training suites of a fire department. Finally, the comprehensiveness of Karo's capabilities has been verified by drawing quantitative comparisons between Karo's performance and other leading robots participating in RRL.

## Introduction

Natural and manmade disasters have caused heavy casualties and significant economic damages over the past few decades. As the impacts of catastrophes are increasing, the necessity of rescue robotics, which explores solutions to minimize the casualties for all phases of a disaster, has increased as well [1]. Due to peerless capabilities of rescue robots compared to human and canine abilities, they are mainly expected to play influential roles in urban search and rescue (USAR) [2], nuclear field emergency operations [3], and mine rescue missions [4]. Practically speaking, a rescue robot needs to demonstrate mobility, dexterity, reconnaissance and exploration capabilities to efficaciously accomplish an USAR mission [5]. These principal capabilities practically empower rescue robots to overcome obstacles, remove rubbles, explore and map unknown environments, detect victims, and deliver essential objects, such as bottle of water, food and drug, to victims' locations. Although the minimum required levels of those capabilities vary in a rescue robot depending on its application, embedding a standard level of those complementary capabilities in a single system

---
*Correspondence: habibian@vt.edu
[†]Soheil Habibian and Mehdi Dadvar contributed equally in this paper
[1] The Authors Were With Advanced Mobile Robotics Lab, Qazvin Azad University, Qazvin, Iran
Full list of author information is available at the end of the article poses various critical challenges which is the primary motivation of this work.

The very first academic attempts in the field of rescue robotics have been done by two groups at Kobe University in Japan [6] and the Colorado School of Mines in the United States motivated by Kobe earthquake and the bombing of Murrah Federal Building in Oklahoma City respectively. However, the mobile platforms introduced in [7] and [8] can also be known as two early experimental efforts in design and implementation of mobile platforms which were potentially suited for USAR, though the authors did not explicitly mention their USAR applications. Later on, rescue robotics started getting more attentions gradually among the mobile robotics research community [9], which led to developing rescue robots with a variety of locomotion mechanisms such as spherical [10, 11], legged [12], wheeled [13], and tracked [14] locomotion.

Track-based locomotion mechanisms provide more versatilities with lower levels of complexities compared to other locomotion mechanisms [15]. In this regard, different attempts have been done on proposing various designs for tracked rescue robots in different sizes, as [5] categorizes them into man-packable, man-portable and maxi-sized types. On this subject, [16] and [17] present two man-packable tracked rescue robot designs. The former develops a paradigmatic system called inspection robotic system to demonstrate the feasibility of mechatronic solutions for inspection operations. The latter proposes the design and development of a novel reconfigurable hybrid wheel-track mobile robot which is constructed based on a Watt II six-bar linkage. Furthermore, [18] demonstrate the application of Quince, a man-packable rescue robot, in an emergency response to the nuclear accident at the Fukushima Daiichi nuclear power plants. Although in [16–18] different man-packable rescue robots with various levels of mobility are presented, man-portable and maxi-sized rescue robots perform higher degrees of mobility and stability while facing more complex obstacles and are well-suited to be equipped with a variety of assistive mechanisms and accessory devices. For instance, [19] develops a tracked man-portable rescue robot that is not equipped with assistive mobility mechanisms such as flippers, while the surveyor type robot implemented in [20] and the Packbot-like main rescue robot presented in are armed with two front flippers which empower them to overcome more elevated and obstacles stably. Moreover, there are some works introducing 4-flipper mechanical designs such as [21] and [22] which have heightened the robots' mobility skills to more sophisticated extent. All mentioned instances of man-portable and maxi-sized tracked rescue robots are focused on mobility and maneuverability capabilities and neglect dexterity and manipulation skills which have resulted in the elimination of the systems' functionality.

Mobile manipulators enhance the dexterity capabilities of rescue robots by enabling them to open/close doors and valves, remove rubbles and obscurations, inspect the environment, and interact with victims. Embedding dexterity capabilities into a highly mobile rescue robot raises serious design and implementation challenges. In this project, [23] presents a mobile robot design based on hybridization of the mobile platform and manipulator. Although this hybridization empowers the robot in term of mobility capabilities, the manipulator itself hardly satisfies minimum dexterity requirements of a dexterous rescue robot. In another effort, a prototype of a 4-flipper mobile platform is presented in [24] that has been integrated with a commercial 5-DOF (degrees of freedom) robot arm. Although this integration has resulted in standard dexterity capabilities of the robot, the non-tracked design of the robot's chassis is still a considerable issue while facing unstructured obstacles. Next, [25] introduces a ground mobile robot which has two parallel tracks and is equipped with a heavy-duty manipulator. Similarly, this work does not present a standard embedding of mobility and dexterity capabilities due to the lack of assistive mobility mechanisms such as flippers which restrains robot's mobility skills. The problem gets even more complicated when robots presented in [23–25] lack reconnaissance and exploration aspects required for a rescue robot which can be a principal shortcoming in situations where data acquisition is a prerequisite for performing maneuvering or dexterity tasks.

Although there are works addressing the exploration and reconnaissance capabilities of rescue robots, little attention has devoted to embedding all three principal capabilities of rescue robots, i.e. mobility, dexterity, and exploration. For instance, [26] presents a highly mobile tracked rescue robot equipped with four flippers which are empowered to accomplish exploration and dose measurement tasks. However, the manipulation capabilities of the robot are only limited to sampling functions by way of a simple 2-DOF manipulator which is an inadequacy for many dexterity operations. Similarly, the rescue robot design introduced by [27] satisfies the basic required mobility skills and also proposes a frontier-selection algorithm for robot's exploration, while the system is deficient in dexterity operations. Consequently, there is a critical lack of attention paid to embedding all three complementary required capabilities in a single rescue robot for accomplishing an effectual USAR mission.

The deep-seated objective of the Karo project is the design and implementation of a maxi-sized rescue robot that meets sophisticated levels of all maneuvering, mobility, dexterity, and exploration requirements delineated in the rulebook of RoboCup rescue robot league (RRL) [28]. To that end, we first elicited the system requirements out of RRL's framework inspired by standard test methods developed by the U.S. Department of Homeland Security, Science and Technology Directorate (DHS S&T) conjunct with the National Institute of Standard and Technology (NIST). This discussion led to the conceptual design of Karo by drafting a locomotion and manipulation system. Considering the proposed conceptual design, this work presents comprehensive design processes along with detail mechanical design of the robot's platform and its 7-DOF manipulator. Further, we present the design and implementation of the command and control system by discussing the overview of the system, robot's power system, sensors, and hardware systems. In conjunction with this, this work elucidates the way that Karo's software system and human–robot interface (HRI) are implemented and employed. Furthermore, we undertook extensive evaluations of Karo's field performance to investigate whether the principal objective of this work has been satisfied. Experimental results of Karo's evaluation in both RRL's test suites and the training suites of a fire department demonstrate that it has effectively accomplished assigned rescue operations utilizing all aspects of its capabilities. Besides, the comprehensiveness of Karo's capabilities has been verified by drawing quantitative comparisons between Karo's performance and other leading robots participating in RRL.

In this paper, we make the following contributions: (1) we propose system specifications for a response robot (Karo) that satisfies requirements and human operators' expectations in USAR missions. These configurations are inspired by NIST standard test methods, and we here employ a holistic combination of their standards when designing our system architecture, (2) we evaluate various conceptual hardware designs for locomotion and manipulation that were tested in our previous works, and point out properties of each design. We develop an optimal conceptual design by incorporating the strengths of the previous versions, which maximizes the functionality of the robot in field performances, (3) we propose a simplified and effective approach for design and analysis of the teleoperated system (i.e., electrical/mechanical hardware, control and command, and human–robot interface) that leverages customized designs to meet the standardized requirements of response robots in one system, and (4) We empirically show that our approach in the design and implementation of Karo, led to a mobile robot which exhibits a high degree of mobility at the side of maintaining required dexterity and exploration capabilities for USAR missions. The results of Karo's performance at RoboCup, demonstrate its comprehensive operation qualities that are expected from a response robot, which confirms our initial design criteria and proposed analysis, (5) We ultimately show—based on Karo's performance at the training field of a fire department—that nevertheless standardized test methods are substantially useful for evaluating response robots and inspire developers, these methods should be extended more to real-life missions. Overall, this work is a step towards preparing highly capable response robots for USAR missions.

The remainder of this paper is organized as follows: the prior works of the authors are presented in "Prior works: Karo's ancestors" section. In "A discussion on conceptual design" section, we provide a discussion on the conceptual design, followed by mechanical design in "Mechanical design" section. The elaborations of the command and control system and software system are discussed in "Control and command system" and "Software systems" sections respectively. Then we propose the system specifications in "System specifications" section. Experimental results and analyzing the robot's field performance are carried out in "Experiments" section. Finally, a conclusion is drawn from the paper's discussions in "Conclusion" section.

## Prior works: Karo's ancestors

During the last decade, several efforts have been made in developing practical rescue robots at Advanced Mobile Robotics Lab (AMRL). Lessons learned from the experimental and field performance of those developed systems frame the roadmap to a more effectual and pragmatic system design. Hence, we briefly review four distinct versions of rescue robots developed at AMRL.

Figure 1a demonstrates the most primitive rescue robot developed at AMRL in 2003 called "NAJI-I", which has been designed to perform basic maneuvering tasks. Its locomotion mechanism is track-based, and it is not equipped with any assistive mechanisms such as flippers. This 35-kg robot can speed up to 15 cm/s at maximum, where it is actuated by two DC servomotors. According to the records of the robot's field performance in Bam earthquake, its locomotion design was an inadequacy for overcoming the haphazardly formed obstacles in the disaster site. Since maneuvering capabilities are prerequisite for all other capabilities of a rescue robot, upgrading the robot's locomotion mechanism design was a crucial step to take.

Field performance results of "NAJI-I" led to an enhanced design for a second version of rescue robot at AMRL in 2004, as shown in Fig. 1b. In the second version, called "NAJI-II", the robot's chassis is equipped with two coupled flippers where their 360-degree rotation helps the robot to maintain its stability while climbing up inclined planes and stairs and enables robot to prevail over the unstructured obstacles. Moreover, the robot's maximum velocity and weight have increased to 25 cm/s and 45 kg respectively, where three DC servomotors function as robot's actuators. Although "NAJI-II" could accomplish a sort of basic mobility tasks such as climbing up the stairs and inclined planes, its stability and reliability was still questionable while overcoming more complicated obstacles.

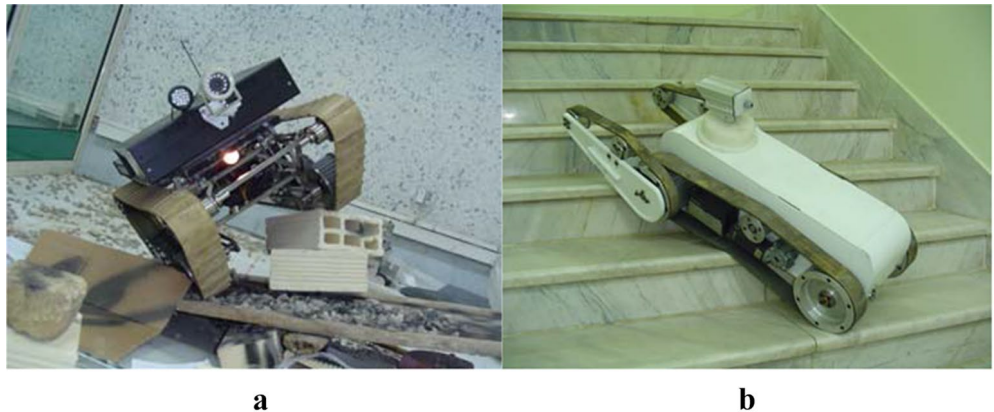

**Fig. 1** Two of Karo's ancestors: **a** NAJI-I in a field performance in 2003, and **b** NAJI-II under test and development procedure in 2004

According to the experimental results of the first two rescue robots developed at AMRL, the essence of a four-flipper design for the rescue robot was justifiable to accomplish sophisticated rescue mission in an uneven area. That is mainly because of two reasons: (1) the distribution, complication and elevation of obstacles vary from one case to another. Therefore, a practical rescue robot should be mobile enough to cope with various forms of obstacles to accomplish the assigned mission. And (2) the mobility itself does not suffice to accomplish a rescue mission successfully because a rescue robot basically needs to maintain its stability while maneuvering in a disaster site to be able to perform other required operations. Besides, lack of a functional robotic arm for accomplishing manipulation and inspection tasks was a considerable deficiency for accomplishing any assigned USAR mission. As a result, later designs evolved to 4-flipper style designs considering a robotic manipulator for robot's dexterity purposes.

Figure 2a illustrates previously developed rescue robot NAJI-VII as it is overcoming a pipe step in Robo-Cup competition 2009. NAJI-VII is equipped with two front triangular flippers which can rotate 360 degrees. Its mechanical design also consists of two regular rear flippers with a course of 90 degrees. The triangular

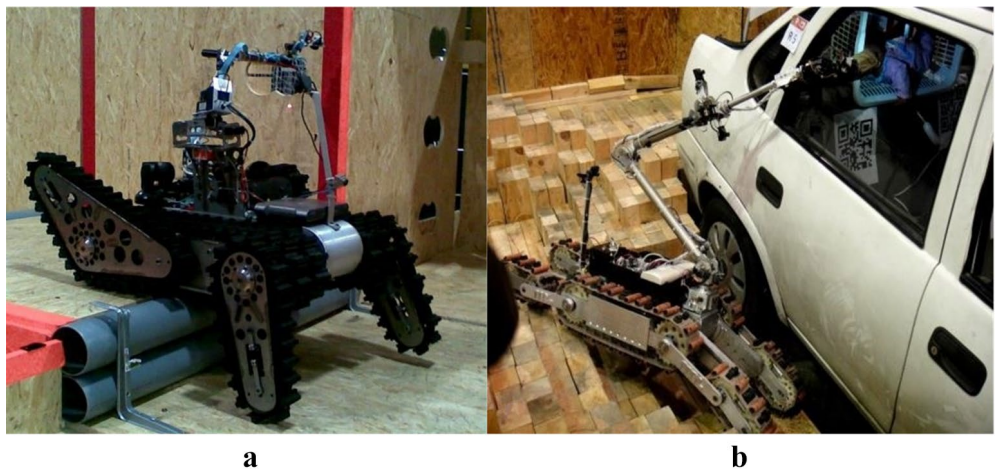

**Fig. 2** Two more ancestors of Karo: **a** illustrates NAJI-VII while overcoming a pipe step in RoboCup competitions 2009, Austria, and **b** explains Scorpion as inspecting inside a car in RoboCup competitions 2012, Mexico

design of front flippers helps them to leverage the robot's body by fewer degrees rotation since the flippers inherently have three-point-contact structures. In despite of previous rescue robots developed at AMRL prior to NAJI-VII, it benefits from a 2-DOF inspection arm. There are mainly two drawbacks associated with this robot's design: (1) although the triangular design of front flippers brings some advantages to robot's mobility, the front flippers cannot be long enough because of the inherence of triangular shapes and the design constraints. This decreases robot's competency to overcome more elevated steps and obstacles. Moreover, the asymmetric design of front and rare flippers makes robot operator's job more complicated. And (2) rescue robots are required to inspect alcoves and holes for victim detection and object recognition. Therefore, the 2-DOF design of the robot's arm does not suffice for the required inspection operations. Furthermore, this arm is incapable of object manipulation and performing a sort of dexterity tasks.

Scorpion is the last rescue robot that we aim to portray as Karo's ancestor. It benefits from a 4-flipper mechanical design where each of its flippers has two links, as shown in Fig. 2b. Considering, the second link of each flipper has a self-relative rotation with respect to the first link. This property helps robot to have more flexibility in rough terrain and enables it to overcome more elevated steps. Moreover, the flippers have been placed along with the robot's chassis which facilitates robot's maneuvering in narrow corridors. However, it made the robot more likable to overturn in uneven terrains. Furthermore, Scorpion derives profit from a 6-DOF manipulator for accomplishing inspection and dexterity operations. Although its 6-DOF manipulator was a decent step for getting inspection tasks done, its limited payload (< 1 kg) restricts its dexterity skills significantly. In addition, Scorpion can speed up to 50 cm/s which is more adequate compared to previous designs, but it is still an operational deficiency.

Alongside mobility and dexterity discussions about the robots mentioned above, their durability and reliability are another subject matter to analyze. According to the carried-out experiments, all these versions are short in reliability for fulfilling a USAR mission in which the reliability is defined as steadiness functionality for repetitive operations. Moreover, the performances of these robots have not been extensively investigated in a real disaster field. Consequently, for instance, their slow set-up and break-down time exemplify a non-practical attribute. As a final point, not only we need to remedy the previous versions' deficiencies for the new design, but also, we must quantitively evaluate the new robot's field performance in a structured manner.

## A discussion on conceptual design
### Mission description
All versions of rescue robots developed at AMRL has been tested in RRL which is held annually since 2001 inspired by the Kobe earthquake [29]. As a matter of fact, taking part in RRL and sticking to its developing frameworks facilitated the advancement and improvement of AMRL's rescue robots. To investigate the rationale behind that, we discuss three main reasons:

1) RRL basically provides objective performance evaluations of rescue robots functioning in simulated earthquake environments. The frameworks of RRL motivate participating rescue robots to demonstrate their skills in maneuvering, mobility, dexterity and exploration test suits [28], which necessarily enforce teams to focus on all aspects of required skills while developing rescue robots.
2) Taking part in RRL gives the opportunity of comparing our developed rescue robot with several other works from all over the world. In fact, not only the inherence of the competition galvanizes the adequate development of rescue robots, but also promotes collaboration between researchers from different teams by sharing experiences during RRL's annual events.
3) The rules and structures of RRL follows the standard test methods developed by the DHS S&T conjunct with the NIST. Thus, RRL's framework is being updated regularly and its evaluations are getting more effective constantly as the NIST's standard test methods get developed. Having said that, the results of rescue robots taking part in RRL aptly reflect their actual potential skills in practice with respect to the latest market demands.

The RRL as part of the International RoboCup competitions provides a benchmark comparison for robot implementers and test administrators based on ASTM's test methods. This benchmark is grouped into four major categories: Maneuvering, Mobility, Dexterity, and Exploration, that each includes various test scenarios (Fig. 3). The Maneuvering suite includes apparatuses with simple terrains: flat surfaces (MAN 1: Center), bridges (MAN 2: Align), an inclined 30° surface (MAN 3: Traverse), 15° continuous ramps (MAN 4: Crossover), movable vertical and diagonal sticks (MAN 5: Negotiate), and ground with bars (MAN 6: Curb), that are embedded in each test for forward and reverse driving orientation. The Mobility suite verifies the capability of robots to pass through apparatuses with medium to hard obstacles such as stacked rolling pipes (MOB 1: Hurdles), 15° surfaces with granular materials (MOB 2: sand/gravel hills), square step field pallets (MOB 3: Stepfields), diagonal hills (MOB

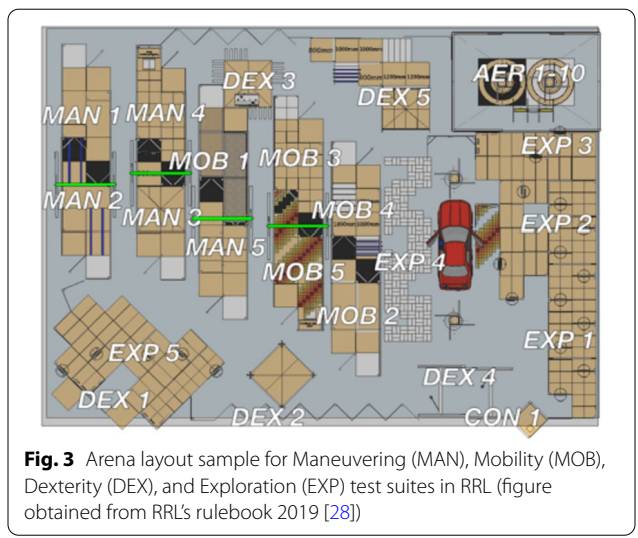

**Fig. 3** Arena layout sample for Maneuvering (MAN), Mobility (MOB), Dexterity (DEX), and Exploration (EXP) test suites in RRL (figure obtained from RRL's rulebook 2019 [28])

4: Elevated Ramps), partly blocked stairs (MOB 5: Stair Debris). For Dexterity suite, objects such as pipes, doors, and wooden blocks are considered for manipulation and inspection tasks. The robot should inspect, touch, rotate, and extract pipes that are placed in various arrangements namely DEX 1: Parallel Pipes, DEX 2: Omni-Directional Pipes, and DEX 3: Cylindrical Pipes. For the case of doors (DEX 4: Door Opening), sometimes equipped with spring closures, the robot should open doors with lever handles and successfully enter the room. Additionally, this suite involves building vertical shoring structures (DEX 5: Shoring) with wooden blocks on a flat surface. The Exploration suite has apparatuses for mapping a dark labyrinth (EXP 1: Map on Continuous Ramps and EXP 2: Map on Crossing Ramps), recognition of objects (EXP 3: Recognize Objects), or detection of obstacles and terrains (EXP 4: Avoid Holes and EXP 5: Avoid Terrains). These apparatuses include terrains with continuous/crossing ramps, amorphous negative obstacles, or terrains that should be avoided. In each test, the operator's intervention is examined by the test administrator that expects the driver to operate the robot remotely for as many as iteration possible in a limited time. Lastly, classified results are presented at the end of the competition to participants. By the means of these performance evaluations, developers can compare their hardware and performance among other participants.

Regarding the above justifications, we aimed to consider RRL as the determined mission for the research and development phase of the proposed rescue robot. Accordingly, we need to first scrutinize the system requirements for success in the determined mission considering the RRL's framework, which is rooted in NIST standard test methods, and then propose a satisfactory conceptual design on the robot's mechanical configuration.

## System requirements

In this section, we aim to set an international standard as benchmark to define the basic requirement for our system. The development of robots has been always a laborious task that forced many robot owners to retreat from using them. In a market, filled with well-advanced robotic platforms, exploring the best option is hardly feasible for users. Thus, there is a big gap between the user's demand and developer's approach in design. In 2005, a comprehensive suite of standard test methods was developed by the U.S. DHS S&T conjunct with the NIST to compare response robots. The DHS-NIST-ASTM (American Society for Testing and Materials) international standard test methods describe the key features of a response robot more broadly:

- Rapidly deployed.
- Remotely operated from an appropriate standoff.
- Mobile in complex environments.
- Sufficiently hardened against harsh environments.
- Reliable and field serviceable.
- Durable or cost-effectively disposable.
- Equipped with operational safeguards.

The above characteristics can be abstracted in maneuvering, mobility, manipulation, sensing, endurance, radio communication, durability, logistics, and safety (see Figure S3 in [30]). The test methods designed by DHS-NIST-ASTM seeks to address these capabilities for various types of robots: Ground, Aquatic, and Aerial, which provide a quantitative method to evaluate the performance of robots in particular missions. The ASTM committee has developed a test apparatus with terrains, targets, and tasks for each mission (Figure S12 in [30]) that robots are expected to operate safely in a limited time. The test suites and evaluation metric get updated every year to improve the evaluation quality, so robot developers and users can compare the operational requirements of each robot in a specific capability spectrum. As a result, having access to a concrete demonstration of operation for a response robot saves a substantial amount of resources before selecting an operating system. Given this very thorough description of the desired respond robot, we corroborated our system by setting a comprehensive operational requirement. Table 1 presents a detailed description of our robot configuration according to ASTM's benchmark.

## Conceptual design

By holding our mission in "A discussion on conceptual design" section on one hand, and the desired performance requirements in "System requirements" section on the other hand, we discuss the design of our system conceptually. This section focuses on drafting the initial locomotion and manipulation system for Karo by analyzing various cases. In the end, a practical concept is chosen by considering eliminated weaknesses of Karo's ancestors (see Sect. 2) and inspirations from outstanding features of contemporary similar platforms [19–22].

**Table 1 System requirements of Karo**

| | | | |
|---|---|---|---|
| Power | Endurance: continuous ramps | Distance (m) per charge | 2000 |
| | | Time (min) per charge | 60 |
| | | Battery life cycle | 300 |
| Mobility | Terrain: Flat Surface | Capability to repeat a 100-m path for 10 times (Y/N) | Y |
| | | Average time per repetition (s) | 200 |
| | Obstacle: Inclined Plane | Maximum incline (°) with 10 repetitions for vertical, diagonal, and horizontal paths | 40 |
| | | Average time per repetition (s) | 30 |
| | Gap Crossing | Maximum gap (cm) traversed for 10 repetitions | 45 |
| | | Average time per repetition (s) | 15 |
| | Stair Climbing | Maximum successful incline (°) for 10 repetitions | 45 |
| | | Average time per repetition (s) | 15 |
| Manipulation | Maximum height of reaching space (m) | | 1.7 |
| | Payload (kg) | | 8 |
| | Door opening capability (Y/N) | | Y |
| Sensor | Visual | Colored video (Y/N) | Y |
| | | Near (40 cm) field acuity (Y/N) | Y |
| | | Far (6 m) field acuity (Y/N) | N |
| | | Field of view (°) | 35*75 |
| | | Resolution (ppi) | 1024*768 |
| | Audio | Full/half-duplex communication (Y/N) | Y |
| | Localization and Mapping | Capability to generate map in a maze with flat ground (Y/N) | Y |
| | | Capability to generate map in a maze with uneven ground (Y/N) | Y |
| Radio Communication | Maximum distance in Line-of-Sight (m) | | 1500 |
| | Maximum distance in Non-Line-of-Sight (m) | | 800 |
| Human-System Interaction | Interface | | Xbox 360 joystick |
| | Control scheme | | Tele-operated |

Figure 3 shows the side view of a conventional tracked response robot and its variations of mobility with several common locomotion configurations. The simplest tracked robot such NAJI-I without flippers (Fig. 4a) barely overcomes obstacles that are higher than the robot's height. This issue led the designers to think of embedding assistive arm mechanisms to improve the maneuverability, NAJI-II was designed the same way. However, using a pair of flippers helps the platform to conquer the steps in cases (b) and (c) of Fig. 4, keeping the horizontal balance of the robot remains still challenging, especially in the case with triangular flippers. This issue makes the operator incapable of controlling the location of the center of mass (COM) that may cause backflipping during climbing. Besides, the triangular flippers are aimed to work similar to wheels but with only three points of contact, and usually lack enough length to work as a lever.

Adopting two pairs of flippers as illustrated in Fig. 4d–f provides the operator with substantial flexibility to adjust the horizontal orientation to avoid backflipping. This becomes particularly crucial when the robot moves on uneven terrains and certain view angles are required for inspection or object manipulation. Additionally, employing more tracked linkages increases the traction capabilities since the robot creates more contact with terrains. Although four flippers enhance the mobility of the robot, our testing proved that having intermediate linkages without tracks [Fig. 4d] limits the locomotion. Mainly because of the gap between the main body and flippers which does not have any tracks. We frequently noticed this issue with Scorpion during tests, and sometimes small, trapped objects between flippers and the body

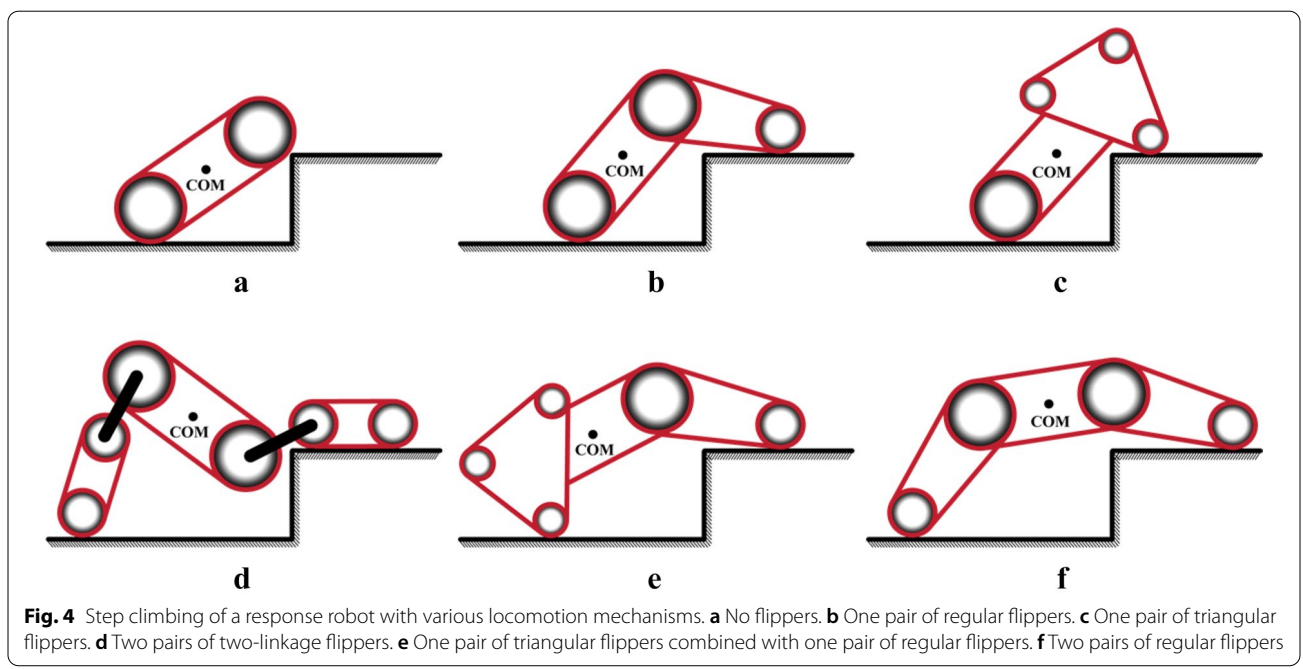

**Fig. 4** Step climbing of a response robot with various locomotion mechanisms. **a** No flippers. **b** One pair of regular flippers. **c** One pair of triangular flippers. **d** Two pairs of two-linkage flippers. **e** One pair of triangular flippers combined with one pair of regular flippers. **f** Two pairs of regular flippers

physically locked the system. On the other hand, asymmetric flipper design of NAJI-VII that is shown in Fig. 4d did not provide all the benefits of the four-flipper design: (1) dissimilar reachability of flippers, (2) uneven contact points of flippers in a 360° of rotation, and (3) restricted height adjustment capability of the robot.

Note that the functionality of each case depends on several factors such as size, weight, location of COM, flipper's geometry, and the obstacle. For example, the step in Fig. 4 is only one example of many complex barriers that exist in the real world. Harsh environments may contain gaps, irregularly shaped barriers, and more importantly, various surface materials. In a similar manner, the dimensions/shapes of linkages also play a major role in the performance. The configuration with no flippers could perform better compared to the four-flipper design which is smaller in size. It is also important to remember that the size and the weight of a robot are directly related and restrict designers to exceed a certain benchmark. Imagine we increase the body size to improve obstacle negotiation: the larger the robot's dimension is the more powerful actuators and power sources we need for the system, which not only makes it pricey but also challenging for us to place parts. This leads to a tradeoff between robot size and the overall locomotion performance. Our prior work with various robot's overall dimensions indicates that larger robot size is not necessarily the game changer in real field performance. For instance, NAJI-VII was limited to entering narrow tunnels or turning in mazes because of the size, while it was capable of overcoming obstacles. On the contrary, Scorpion experienced a lack of mobility in gap crossing due to a smaller size. Accordingly, we selected the four-flipper design [Fig. 4f], with an optimal optimum robot size, as an appropriate locomotion mechanism that efficiently can satisfy the technical requirements of Table 1. The detailed description of the mechanical design and analysis for the mobile platform is presented in "Locomotion mechanism" section.

Manipulation capability is also an essential element for response robots which was absent in the primary versions. Later, the reconnaissance and exploration needs grew over time and urged scientists to use camera arms to provide more flexibility in the inspection. Nowadays, most of the rescue robots are equipped with manipulators which are not only capable of object manipulation but also have devices such as camera, temperature sensor, carbon dioxide sensor, etc. Thus, the new generation of response robots has complex manipulation systems that often makes the operation more difficult. From DOF to joint arrangements, how can a manipulation system be flexible and dexterous [31, 32] enough for the requirements of a response robot ("System requirements" section)? By defining applications and work environments, how can we design the workspace sufficient enough for all tasks? We aim to create a conceptual design to work towards answering these questions for the desired manipulator of a rescue robot.

According to RRL's Dexterity test scenarios, a response robot should be able to search victims while performing dexterous manipulation tasks in various locations in the field. These embeddable tests, usually made of PVC pipes (see Dexterity section in RRL's rulebook [28]), are repeatable and reproducible test packages. They can be mounted on the planer and cylindrical surfaces at different heights/angles, or even be placed in hardly accessible locations such as inside vehicles, packages, and narrow gaps. The tests can be categorized as four different manipulation tasks: inspection, disruptor aiming, object insertion, and object retrieval, it may also include lifting heavy objects, opening doors, cutting, and unlatching. Thus, a minimum DOF with a suitable joint arrangement is required for the manipulator. Considering various operational scenarios in a field of disaster "A discussion on conceptual design" section and performance requirements in "System requirements" section, we developed a conceptual design of a 7-DOF robotic arm (Fig. 5) with a specific order of joints. Figure 5 shows the kinematic design; the cylinders are rotational joints, and the cubic is the only linear joint. This concept provides enough resilience for the operator to perform various dexterity maneuvers in mentioned operational situations. The first joint rotates the base of the arm relative to the robot's orientation. This feature is crucial in a situation that moving the mobile platform is difficult for the operator. The second and third joints mainly adjust the position of the end-effector in the plane of operation. Further, the linear joint helps aiming, retrieval, and insertion tasks in any position. Also, the last three joints resemble the human hand to perform rotation and bending motions. This link/joint arrangement satisfies the required resiliency in essence. "Manipulation mechanism" section addresses the detailed mechanical design and how the conceptual design delivers insights about the workspace of the manipulator.

## Mechanical design

Based on the former discussion on the conceptual design, almost all components and equipment (915 total and 230 unique parts without counting fasteners) of the robot were designed and modeled in SolidWorks (Fig. 6) for initial assessments. This section elaborates on the procedure employed to mechanically design the system, select the appropriate actuators, analyze the mechanical components, and identify the manipulation workspace. As shown in Fig. 6, all of the electrical devices are modeled as well, and their location was adjusted by using the Collision Detection feature in SolidWorks. Modeling all of the components in 3D not only provided accurate weight estimation but also enabled us to examine the mechanical design of the parts in various arrangements. Additionally, it undoubtedly improved Karo's manufacturing quality done by computer-aided machining. As a result, Karo has better functionality, reliability, and component accessibility compared to its previous versions, which makes it convenient for the setup team to rapidly deploy it before USAR missions.

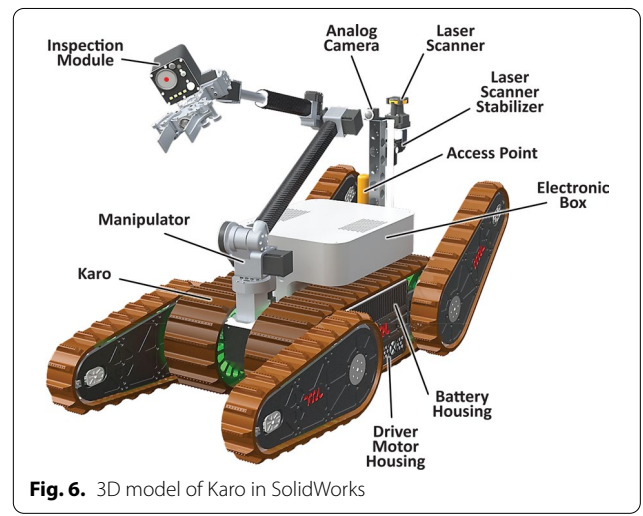

**Fig. 6** 3D model of Karo in SolidWorks

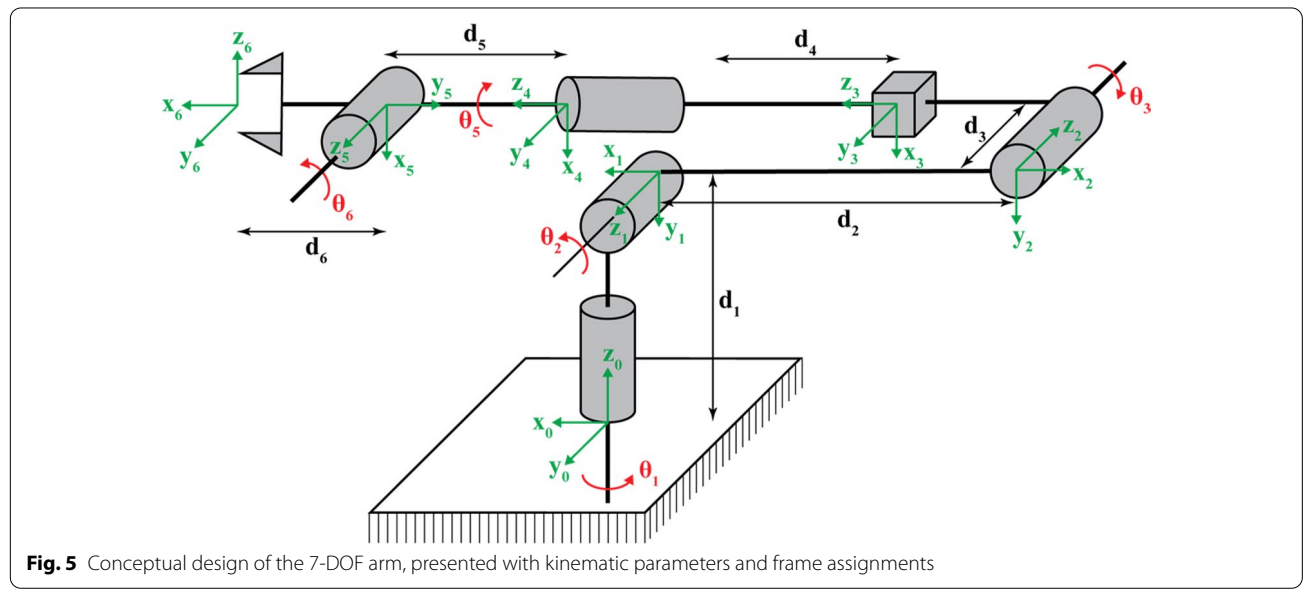

**Fig. 5** Conceptual design of the 7-DOF arm, presented with kinematic parameters and frame assignments

### Locomotion mechanism

*The* mobile platform is designed into three different modules: side chassis, middle chassis, and flipper, which are shown with orange, blue, and green dashed boxes in Fig. 7a, respectively. This centrosymmetric configuration provides a convenient assembly of parts in addition to decreasing the costs of manufacturing. Each module can be taken apart or replaced individually without complete disassembly if any maintenance required, which meets the field serviceable feature of DHS-NIST-ASTM standards. The locomotion system is embodied in 4 DC motors (two for flippers and two for traction), 4 customized gearboxes (worm gears and bevel gears), 8 steel alloy shafts, 16 polyamide pulleys, 6 polyurethane tracks, and 12 aluminum body plates. To increase the stability during negotiation with obstacles, three out of four motor gearboxes (two traction and one flipper motor gearboxes) are placed at front of the body to avoid backflipping. Note that both traction motor gearboxes are located below the main tracks and not shown in Fig. 7a. Each pair of front/rear flippers are driven by one worm gearbox (e.g., the front worm gearbox controls the angle of front flippers at the same time). Similarly, the traction of each side of the body (two flippers and one main track) is independently driven by one bevel gearbox.

Figure 7b depicts the cutaway diagram of the locomotion mechanism located at the front right of the body, it is part of the symmetric design, the same structure exists on the left side. The bevel gearbox drives a hollow shaft that is coupled to pulleys which drive the rotation of all tracks on the right side. The smaller shaft (flipper shaft) that passes through the hollow shaft and coupled with the worm gearbox, rotates the front flipper. One end of the flipper shaft is coupled with a flange (fastened to the flipper's plate) while the other end is linked to an Oldham coupling to accommodate a small amount of axial misalignment caused by tension regulation in tracks. The tension of each track is adjusted using two bolts that set the center distance between pulleys. All the bearing arrangements are designed according to the manufacturer's standards.

### Torque requirements analysis

Following the design of the locomotion mechanism, and after the initial estimation of dimensions and weight of the robot, we can select appropriate actuators and standard components. Since the robot should reliably operate in harsh environments and be able to climb stairs or steep slopes, cross gaps, and even pass uneven terrains [15], it is always prone to experience unidentified external loads. Analyzing the unexpected circumstances is pretty complicated and laborious. Hence, for input power calculation, we only selected the two most challenging movement scenarios that are frequently performed in missions: climbing steep ramps and lifting the robot with flippers. Targeting those cases as the continuous operation benchmark would guarantee that the robot can effortlessly perform regular movements. The following presents analyses with basic assumptions to determine the minimum output torque and power for the system.

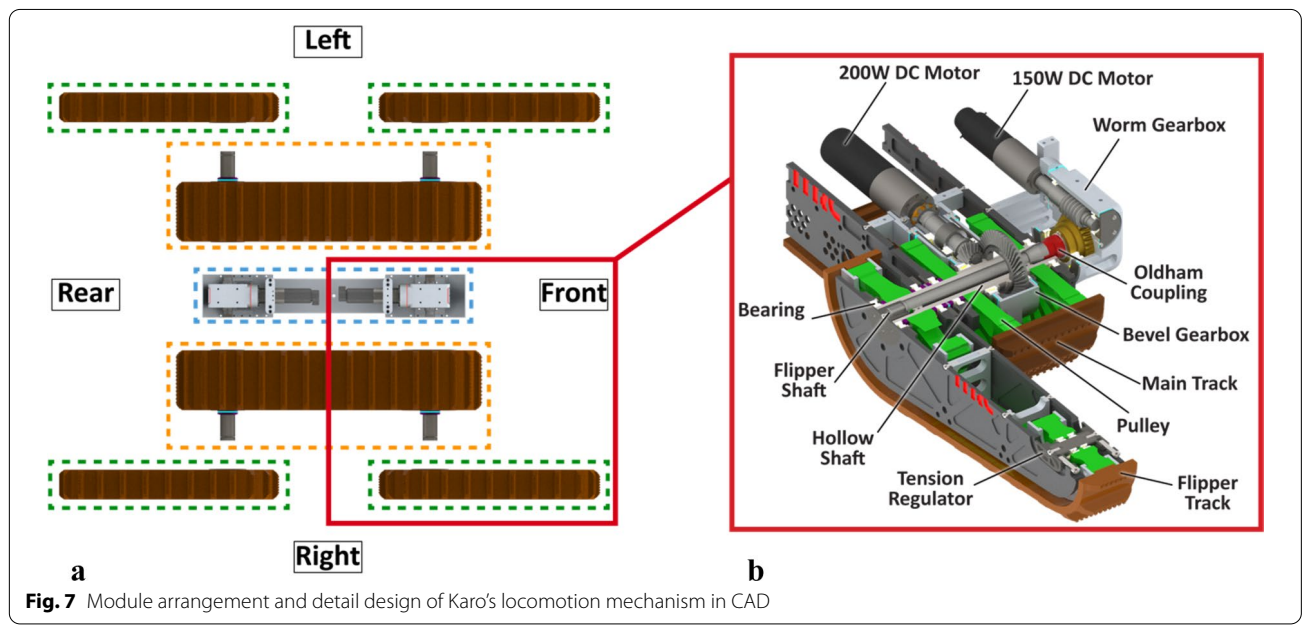

**Fig. 7** Module arrangement and detail design of Karo's locomotion mechanism in CAD

The traction system should overcome various resistant forces when passing haphazardly formed obstacles. Climbing a ramp is one of the most common and challenging mobility tasks that the robot has to overcome during operations. Here, we consider the robot climbing a 40° ramp (left side of Fig. 8), determine the minimum output torque required by each actuator, and select a DC motor available in the market based on the desired speed of the robot. Note that this analysis is intentionally simplified to avoid complicated calculations. For instance, friction between the robot and the environment is neglected due to variations in friction force, the operation site includes fields made by various materials. Additionally, the contact area of tracks and the normal force exerted on the robot could exceptionally alter the friction force, especially since Karo's tracks are made of Polyurethane. Special circumstances of the standard model of friction are usually considered for rubber materials. The friction force in the power transmission system is considered as the power loss in calculations.

Figure 8 (left) presents the free body diagram (FBD) of Karo on a slope where $F_{tr}$ is the total traction force applied by motors, $M$ is the weight of the robot, $N$ is the normal force, and $\alpha$ is the slope angle. Considering no terrain slippage between the tracks and the ramp, the force balance can be written according to Newton's second law to describe motion in $i$ direction Eq. (1). The exerted resistance forces are highly dependable on the terrain types, so any consideration would not be accountable for the actual power loss. Instead, we apply a margin in the calculation to compensate for the effects of all the frictions in the system.

$$F_{tr} - Mg\sin\alpha = Ma_i \tag{1}$$

$$F_{tr} = Mg\sin\alpha \tag{2}$$

where the angular acceleration of the pulley is $\dot{\omega}$ and therefore, the linear acceleration of the body is $a_i$. Writing the general form of Euler's rotation equations with respect to the center of the actuation pulley:

$$T_o - F_{tr}b = I\dot{\omega} + \omega \times (I \times \omega) \tag{3}$$

where $T_o$ is the motors' applied torques, $I$ is the inertia matrix, and $\omega$ is the angular velocity of the pulley. The first term on the right side of Eq. (3) is zero because the system has very small acceleration. The inertia matrix only includes the principal moments of inertia since the center of rotation is aligned with COM of the pulley, so the second term is also zero. Considering all of the actuation torque to be merely transmitted through inextensible components, the following relationship in Eq. (4) can be written:

$$T_o = \frac{F_{tr}b}{2} \tag{4}$$

where $b$ is the center distance of the pulley to the outer surface of the track. For our system $\alpha = 40°$, $m = 85$ kg,

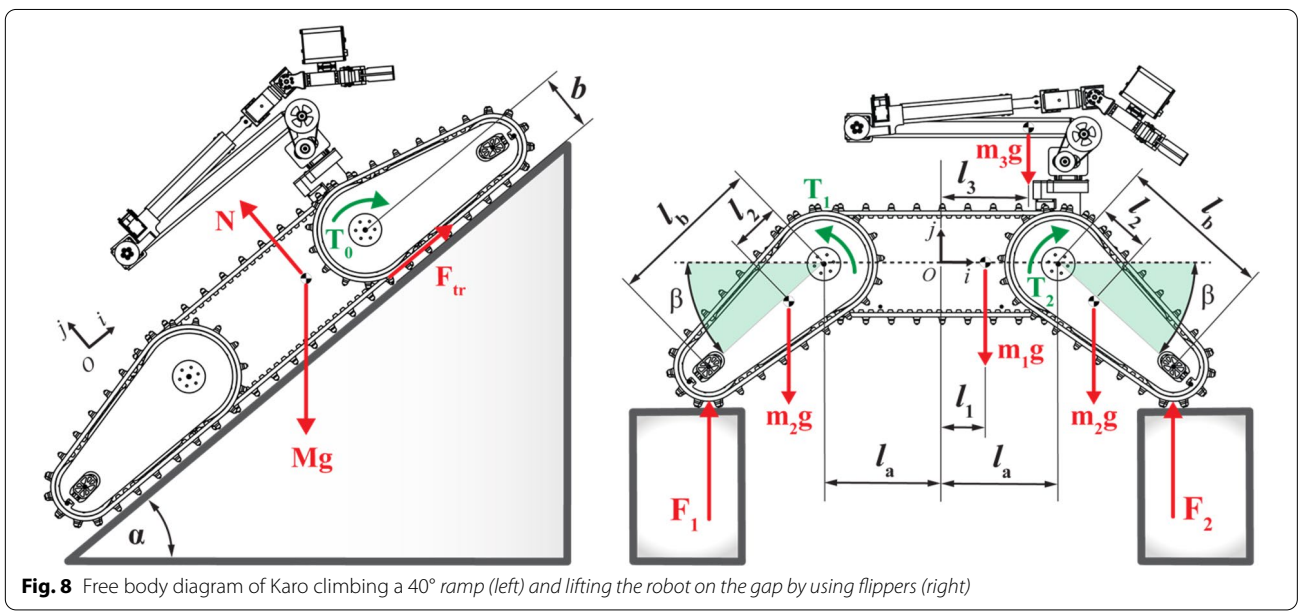

**Fig. 8** Free body diagram of Karo climbing a 40° *ramp (left) and lifting the robot on the gap by using flippers (right)*

and $b = 12$ cm, which requires 64.25 Nm of output torque to enable the robot to climb the ramp continuously. Since two DC motors are considered for the locomotion, 32.12 Nm should be provided by each side chassis motor. By considering the desired locomotion speed to be higher than 0.5 m/s, the RE 50, 200 W Maxon DC motor with the GP 52 C planetary gearhead (26:1 reduction) coupled to the bevel gearbox (4:1 reduction) was selected for each side chassis. The RE 50 motor generates a nominal speed of 5680 rpm and nominal continuous torque of 405 mNm at 24 V. The GP 52 C planetary gearhead [33] and the bevel gearbox [34] cause 17% and 4% power loss respectively, which it makes only 33.6 Nm of the output torque accountable. This amount of torque generated by each motor is sufficient for continuously climbing the 40° ramp with a speed of 0.8 m/s. The operation range of the RE 50 motor suggests that it can provide higher rotation speed when less output torque is required or conversely. Therefore, it ensures that the robot can easily move in complex environments and the motors are still capable of providing more power if any abrupt unexpected situation happened during the operation.

The flippers help the robot to pass obstacles by changing the location of COM and also providing more contacts with terrains. As the angle of each flipper and the position of the manipulator changes, the location of COM shifts proportionally. Hence, one flipper motor could experience a larger external load than the other depending on the situation. Figure 8 (right) illustrates the FBD of Karo on a gap when only flippers are in contact. Separate COMs are assigned for the front/rear flippers, main chassis, and the manipulator, to account for changes in reaction forces based on the position of each part. For this analysis, similar to the traction system, we consider the most challenging condition in which both flippers are completely horizontal and lifting the robot (e.g., β is zero). The position of the manipulator is usually fixed during this motion. The location of the exerted forces are $l_1 = 30$, $l_2 = 150$, $l_3 = 100$, $l_a = 260$, and $l_b = 310$ mm. $F_1$ and $F_2$ are reaction forces applied at the tip of each flipper. Writing the force balance according to Newton's second law in $j$ direction Eq. (5) and the moment balance based on the general form of Euler's rotation equations Eq. (6) with respect to center $O$:

$$F_1 + F_2 - (m_1 + 2m_2 + m_3)g = (m_1 + 2m_2 + m_3)a_i \tag{5}$$

$$m_1 g l_1 + m_3 g l_3 + F_1(l_a + l_b \cos\beta) - F_2(l_a + l_b \cos\beta) - m_2 d_3 = I\ddot{\theta} \tag{6}$$

where $a_i$ and $\ddot{\theta}$ are the linear and angular acceleration of the robot, $I$ is the inertia matrix, and $m_1 = 50$, $m_2 = 13.3$, and $m_3 = 8.5$kg are the mass of the chassis, two of the flippers, and the manipulator, respectively. Considering the accelerations to be equal to zero, the reaction forces are found to be $F_1 = 392.8$ and $F_2 = 441.2$ N when β is zero. To calculate the maximum required torque, the larger reaction force ($F_2$) is selected and therefore, 136.7 Nm of torque is required to rotate the flippers. This case occurs rapidly during the lifting process, and as the flippers' contact point with ground gets closer to the body, the calculated values decrease significantly. However, tests in conditions such as step field pallets show that flippers experience greater loads. For this reason, we considered a larger safety factor to choose the motors for the flippers. The RE 40, 150 W Maxon DC motor [33] with the GP 42 C planetary gearhead (reduction: 43:1) coupled to the worm gearbox (reduction: 30:1) was selected to actuate each pair of the flippers. The RE 40 motor generates a nominal speed of 6040 rpm and a nominal continuous torque of 177 mNm at 24 V. The GP 42 C planetary gearhead and the worm gearbox cause 28% and 15% power loss respectively, which it makes only 151.4 Nm of the output torque accountable. The rotational speed of flippers is 32 deg/s when flippers are in contact with the ground, they rotate faster in the unloaded pace.

The above analysis is aimed to obtain a reasonable ballpark of the power requirements and also verify the design of the locomotion system. We tested Karo for both of the discussed movement scenarios. "Locomotion mechanism experimental results" section represents the experimental measurements of motors' current and torque and discusses their alternations in several sequences of motion.

## Manipulation mechanism

The manipulator is an essential element of Karo that completes the mission when the robot successfully reached victim's location. In this section, we explain the mechanical design, actuator selection, and present the workspace of the manipulator. As depicted in Fig. 9, it is a 7-DOF arm made of rotational and prismatic joints, including the gripper's jaw (Link 7). For all of the joints, off-the-shelf Robotis Dynamixel servo motors were selected because of their high resolution, output torque, control algorithm, and compact design. All of the links are made of carbon fiber or Aluminum to provide a lightweight and rigid structure. One noticeable feature of this arm is operating on a mobile platform that makes the manipulation more challenging compared to the industrial counterparts, because the position and orientation of the arm constantly change the positioning of the end-effector. Link 1 sets the orientation of the whole arm relative to the robot, Links 2 and 3 mainly adjust the height of the end-effector; and the rest of the links provide more flexibility to object manipulation and inspection. Additionally, Link 6 enhances the inspection by rotating the sensor box independently. Similar to the mobile platform, this manipulator is designed in a modular configuration to make the assembly and accessibility convenient for the setup team.

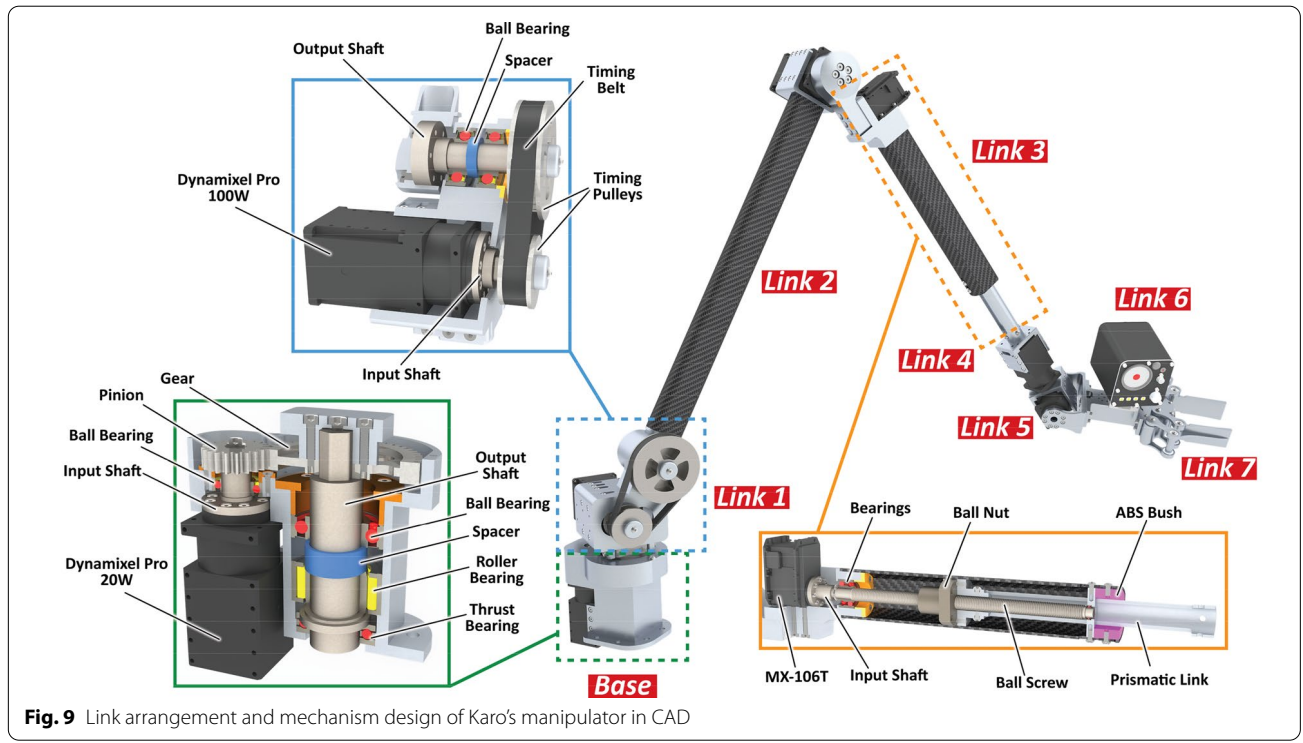

**Fig. 9** Link arrangement and mechanism design of Karo's manipulator in CAD

The blue box in Fig. 9 illustrates the timing belt drive design in Link 1. The servo motor can be fixed at different positions relative to the output shaft to adjust backlash and tension in the timing belt mechanism. Two deep-groove bearings are mounted on the output shaft, apart from each other, to tolerate radial and bending forces on the second link. For Joints 1, 5, 6, and 7, the 20 W Dynamixel Pro is employed, which have a smaller dimension and weight. Joint 1 is coupled to a spur gearset (5:2 reduction) which makes a compact design for the base besides increasing the output torque of the first joint. As depicted in the green box in Fig. 9, a combination of bearings is embedded in the base to absorb shocks, improve rigidity, and provide smooth motion. A needle-roller bearing is positioned apart from a regular deep-groove ball bearing to ensure capturing the radial exerted forces to the main shaft. Additionally, the axial forces are tolerated by the thrust bearing situated at the bottom of the output shaft. Another deep-groove bearing located along the axis of the servo motor's shaft to endure the radial forces produced in the spur gear set. The linear motion of Link 4 is created by the MX-106 T that is coupled to a ball screw, as depicted in the orange box in Fig. 9. The ball nut moves across the ball screw while the flange is constrained between the internal walls of Link 3 and the prismatic link sliding inside ABS bush. These high-efficiency mechanisms combined with the high resolution of servo motors provide precise motion with the minimum backlash for the manipulator.

Figure 9 also illustrates the detail design of Joints 1, 2, and 4 in the green, blue, and orange box, respectively. Since manipulation tasks in rescue missions tend to be slow-paced, we avoided the conventional dynamic analysis of robotic arms and simply used the static form of Euler's equations to calculate the required torque of each joint for motor selection. Considering the fully-extended scenario (Fig. 10) as the continuous operation benchmark, which usually takes place briefly during motion, ensures that the manipulator is capable of operating when links are positioned at other angles.

Here, torque calculation of Joints 2 and 3 are only discussed since they require higher torques, identical calculations were conducted for other joints. The parameters of Fig. 10 are $m_t = 8.5$ kg is the total mass, $m_c = 2.8$ kg is the mass of Links 3–7, and $m_{ex}$ is the external mass. $l_t = 40$ cm is the distance from Joint 1 to COM of the whole arm, $l_c = 45$ cm is the distance from Joint 2 to COM of Links 3–7, $l_{ex} = 130$ cm is the distance from Joint 1 to the external load, and $l_j$ is the length of Link 2. Writing the moment balance based on the general form of Euler's rotation equations with respect to center $O_1$ for Joint 2 Eq. (7) and center $O_2$ for Joint 3 Eq. (8):

$$T_2 - m_t g l_t - m_{ex} g l_{ex} = I\ddot{\theta}_2 \tag{7}$$

$$T_3 - m_c g l_c - m_{ex} g(l_{ex} - l_j) = I..\theta_3 \tag{8}$$

where $\ddot{\theta}_2$ is the angular velocity of links 2–7, $\ddot{\theta}_3$ is the angular velocity of Links 3–7, and $I$ is the inertia matrix. Without existence of an external load, Joint 2 requires 3.4 Nm and Joint 3 requires 1.2 Nm of minimum torque to support the weight of parts. We employed the 100 W Dynamixel-P (PH54-100-S500-R) with the continuous output torque of 25.4 Nm, the output speed of 29.2 rpm, and the resolution of 0.0004 deg/pulse for both Joints 2 and Joint 3. Due to the higher power requirements of Joint 2, we added a timing belt drive (3:1 reduction) to enhance the output torque. This mechanism with the high efficiency of 98% [35] enables the manipulator to lift an external mass $m_{ex} = 5.6$ kg at the fully-extended position by using Joint 2, the payload increases as $l_{ex}$ becomes smaller.

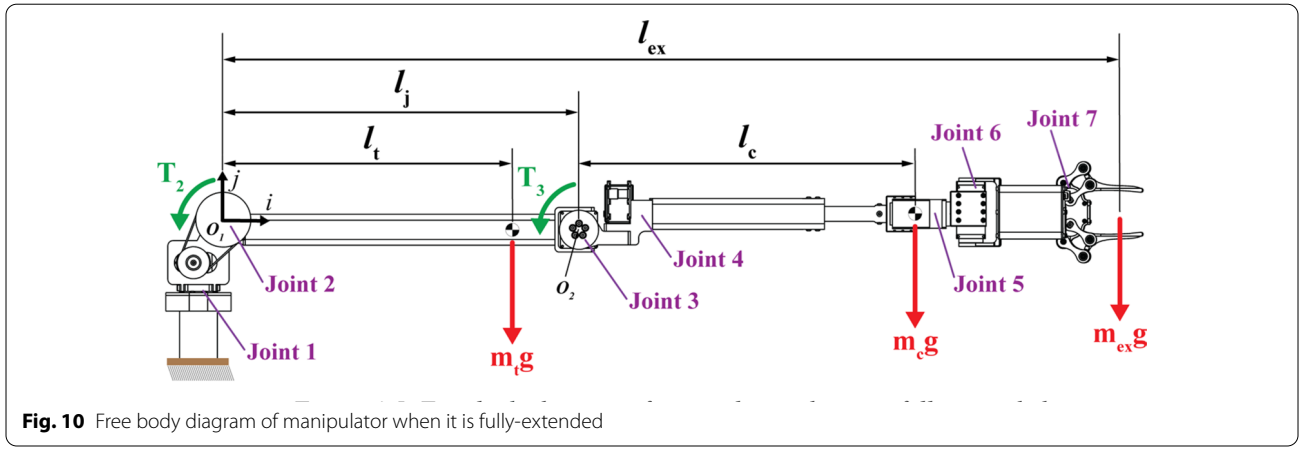

**Fig. 10** Free body diagram of manipulator when it is fully-extended

### Achievable workspace
Performing dexterity tasks is directly related to the locations that the end-effector can reach within an environment. For this manipulator, since it is mounted on a mobile platform, there are some conflicts with the body of the robot, which makes it more important to determine its workspace. In this section, we aim to simulate the workspace of the manipulator by using the Denavit-Hartenberg method, based on the motion range of each joint, to identify those conflicts and achieve the best functionality for the operator. To calculate the transformation matrix and ultimately, positions of the gripper, we need to define link kinematic parameters, known as D-H. Using D_H parameters allows us to find the orientation and position of every join in space with respect to an inertial reference frame. Figure 5 presents the kinematic diagram of the arm and frame selection according to the D-H convention. By using the relationship between two successive local coordinate frames, $i$ and $i+1$ we can obtain the D-H kinematic parameters. According to this method, $r_i$, $\alpha_i$, $d_i$, and $\theta_i$ are required to define the transformation matrix between two coordinate frames next to each other. The transformation matrix, which can calculate the position of a point in $i+1$ coordinate frame with respect to $i$ coordinate frame is given as:

$$A_i = \begin{bmatrix} \cos\theta_i & -\sin\theta_i\cos\alpha_i & \sin\theta_i\sin\alpha_i & r_i\cos\theta_i \\ \sin\theta_i & \cos\theta_i\cos\alpha_i & -\cos\theta_i\sin\alpha_i & r_i\sin\theta_i \\ 0 & \sin\alpha_i & \cos\alpha_i & d_i \\ 0 & 0 & 0 & 1 \end{bmatrix} \tag{9}$$

For our system we need to calculate the above matrix for each degree of freedom to obtain a space-fixed matrix of transformation as follow:

$$T_i = A_1 * A_2 * A_3 * A_4 * A_5 * A_6 \tag{10}$$

The first six joints determine the location of the gripper in space. Each joint has four kinematic parameters; therefore 24 total parameters should be defined to calculate the position and the orientation of the end-effector. Figure 5 shows the coordinate frame assignments and D-H parameters of the arm, which are selected in a fashion to zero some of the parameter and make the computation relatively faster. Table 2 lists these parameters of the manipulator at the home position based on Fig. 5.

By using the geometrical constant parameters and Eq. (8), the position of the end-effector is determined for various sets of joint positions and the workspace is depicted in Fig. 11. The three view orientations of the workspace provide useful information about the accessibility of the arm to the surroundings of the robot, it can reach points within a sphere with a radius of 130 cm which is centered at the base of the manipulator. The left view shows that the arm can access 48 cm below of the robot, however it is restricted to the front half of the body. There is also an unreachable region close to Link 1, which is practically not the area of interest for manipulation. Additionally, the front view shows that the points within 100 cm of the left and right of the robot are accessible. As the top view demonstrates, the operator is able to reach points that are within 92 cm of front flippers and 23 cm of rear flippers. Lastly, there are 10° and 20° slices of the workspace due to the range of motion of Joint 1. In conclusion, by using this method we are able to control the workspace by adjusting the constant parameters or limiting the joints to achieve the desired workspace or avoid singular positions of the mechanism [36]. Moreover, the operator can use the workspace plots generated by forward kinematics study to get some insights into positioning the robot while reaching objects within an environment.

**Table 2  D-H parameters of the first six degrees of freedom**

| Link | Joint Variable | $r_i$ | $\alpha_i$ | $d_i$ | $\theta_i$ | Range |
|------|---------------|-------|------------|-------|------------|-------|
| 1 | $\theta_1$ | 0 | $-90°$ | $d_1$ | $\theta_1$ | $-80°$ to $80°$ |
| 2 | $\theta_3$ | $d_2$ | $-180°$ | 0 | $\theta_3 - 180°$ | $0°$ to $180°$ |
| 3 | $\theta_3$ | 0 | $-90°$ | $d_3$ | $\theta_3 - 90°$ | $0°$ to $180°$ |
| 4 | $d_4$ | 0 | $0°$ | $d_4$ | 0 | 0 m to 0.4 m |
| 5 | $\theta_5$ | 0 | $-90°$ | $d_5$ | $\theta_5$ | $0°$ to $360°$ |
| 6 | $\theta_6$ | $d_6$ | $90°$ | 0 | $\theta_6 - 90°$ | $-90°$ to $90°$ |

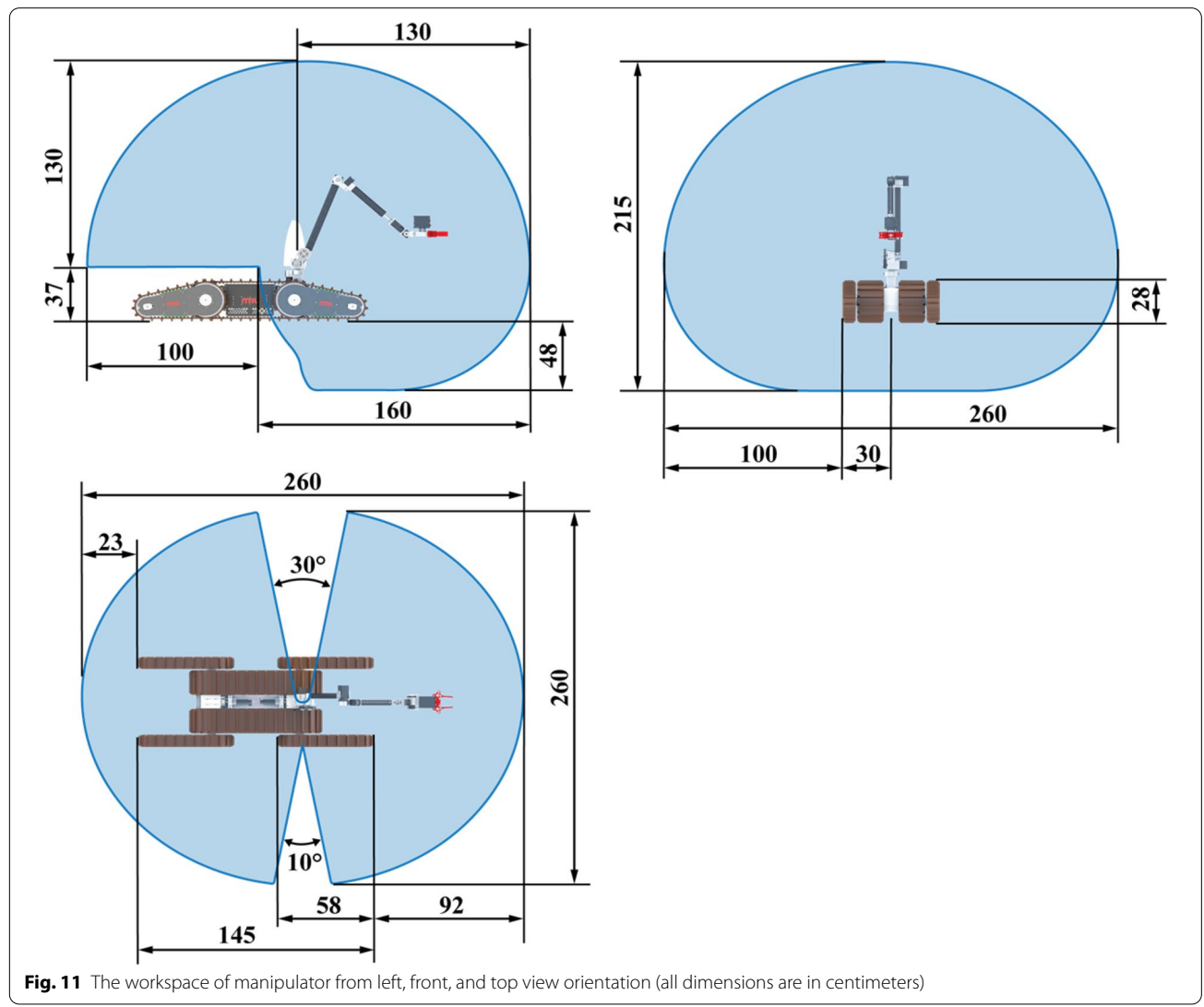

**Fig. 11** The workspace of manipulator from left, front, and top view orientation (all dimensions are in centimeters)

### Component analysis

The previous analysis helped determine the motors, gear ratios, bearings, and revising dimensions of the mechanical parts. The next step is analyzing mechanical components of the robot to ensure that all parts have enough strength while the robot's weight is optimized. Additionally, this analysis enables identifying the critical sections of each part and redesigning it in such a way that stress is not only reduced but also uniformly distributed. For power transmission components such as gears, bearings, and shafts, the stress is calculated using standard mechanism analysis [37] to avoid failure. Major components of the robot are either geometrically sophisticated or they are prone to complex loading patterns, which makes solving stress–strain equations tedious. This matter becomes significant when studying fatigue failure due to cycling loadings under varying circumstances [38, 39]. In order to efficiently investigate those cases and optimize our design, we employed the finite element analysis (FEA) simulation, using the commercially available software Abaqus from Dassault Systèmes, as an additional design and analysis tool. As an example, Fig. 12a illustrates the geometry of the flipper's shaft, imported from SolidWorks, under constraints and loadings in the mechanism. The AISI 1010 carbon steel is used as the material for this part, which is modeled with 13,315 quadratic tetrahedron elements (type C3D10). As shown in Fig. 12c, 5.14 kPa of maximum principal stress is created by an applied static torque of 152 Nm (calculated torque in "Manipulation mechanism" section) creates, which does not reach the yield strength of AISI 1010 steel (305 Mpa). The torsional displacement of the shaft is accordingly negligible as shown in radians in Fig. 12b.

A similar approach was taken for other critical components to make the design more reliable and robust. However not all of the loads on the system can be calculated due to the dynamics of the operation environment. Overall, FEA assisted us to recognize the critical stresses for each component and optimize the geometry, and therefore, helped to reduce the overall weight of the robot, which essentially contributed to less power consumption.

## Control and command system

### Overview of the system

We have proposed a mechanical design for robot's platform and its manipulator in "Mechanical design" section. These designs took the first step towards the objective of this research, which is discussed in "Introduction" section. Thus, we discuss developing Karo from electrical system design point of view now on to take the second needed step, which is the design of the control and command system. Basically, we design a control and command system to (1) effectually control the actuators embedded in the mechanical structure, (2) derive diverse sensors and perception devices to accurately perceive the environment and control the system itself, and (3) build a sophisticated platform for high-level software systems and algorithms to be implemented. As Fig. 13 delineates, we designed a control and command system which mainly is comprised of three modules (1) mobile platform, (2) manipulator, and (3) operator control unit (OCU). Here, we aim to depict an overview of the system from computation and communication perspectives.

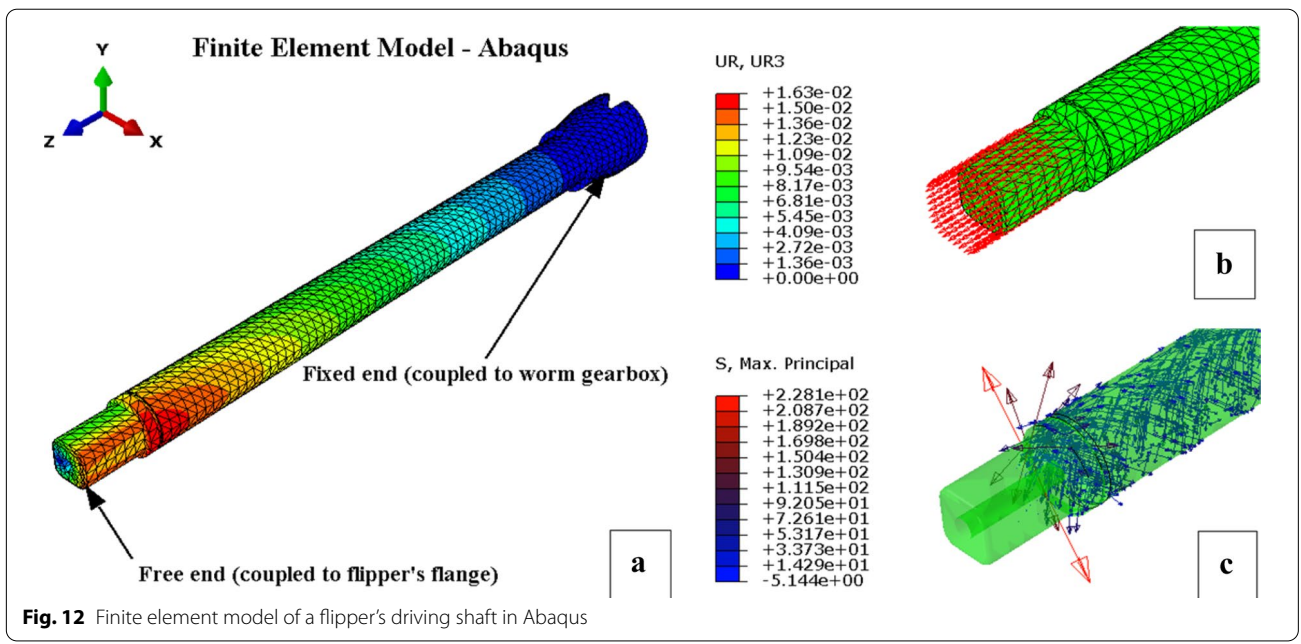

**Fig. 12** Finite element model of a flipper's driving shaft in Abaqus

Computations: Each module deals with several devices, electronic or mechanical components and acquired data by the way of computing units. As Fig. 13 illustrates, OCU retains only one mini pc, that is Intel NUC kit, as its computing unit, while the robot platform module benefits from one similar mini pc for high level processes such as simultaneous localization and mapping (SLAM) and one ARM Cortex-m3-based board to handle low level peripherals and IOs. The manipulator module also profits from two similar ARM-based boards acting as its controller and low-level peripherals handler. Basically, in the explained distribution of computing units, we tried to assign more expensive computations to mini PCs, such as any kind of image processing, while the hardware-involved computations, such as a position-controller, are assigned to ARM-based microcontrollers.

Communications: the communication scheme of the system is highly critical for the stability and quality of the robot's performance, since controlling tele-operated robot, video and sound streaming, system diagnostics, sensors feedback, visualizing procedures and SLAM in a remote station are all relying on the remote communication platform. In this regard, the OCU is equipped with UBIQUITI Networks 802.11a/b/g Bullet M5 Access Point/Bridge where it connects to the identical access point on the robot platform module wirelessly. To ensure the communication stability and robustness in long distances, the access points function on a 500 MW-power basis. All connections between the robot's access point and other computational nodes on the mobile platform and the manipulator module are established throughout Ethernet. Besides, we utilize the UDP communication protocol for transmitting the control commands between OCU and the ARM-based control boards, since such command transmissions are time-sensitive.

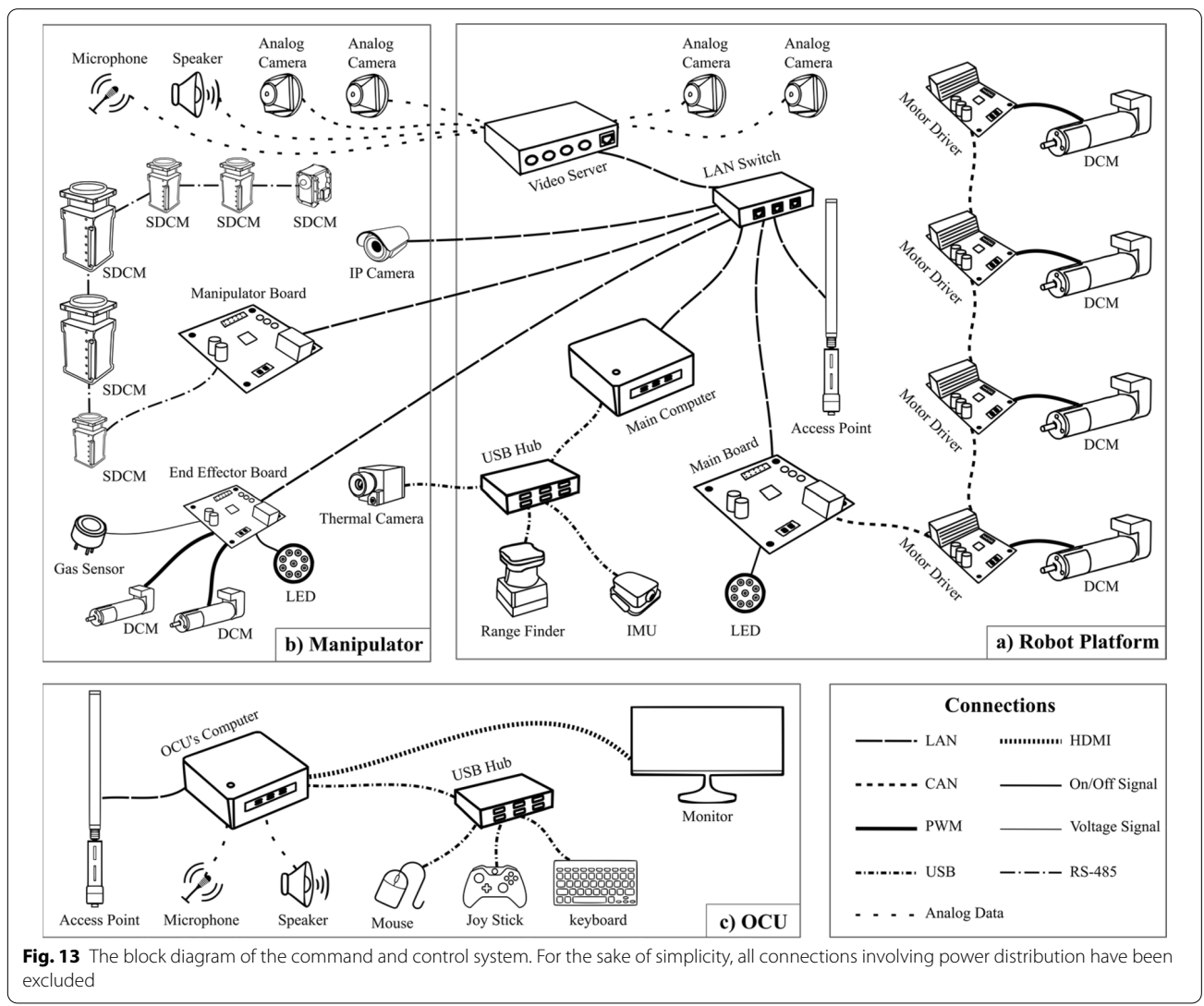

**Fig. 13** The block diagram of the command and control system. For the sake of simplicity, all connections involving power distribution have been excluded

### Sensors

To adequately perform the required tasks, Karo needs to be equipped with a wide variety of sensors such as laser scanner, inertial measurement unit, RGB-D camera, shaft encoder, and thermal camera. This sensory system functions to accurately perceive the environment, in terms of obstacles, victims, signs, and objects. Here, we provide a brief description for each of them:

Shaft Encoder: as explained in "Locomotion mechanism" section, the robot's platform is equipped with two brushed DC motors for its movement and two others for its front and rear flippers. Accordingly, the robot has been set up with four incremental optical rotary shaft encoders, which make wheels odometry calculation possible. The system eventually benefits form this odometry for DC motor control purposes and the localization and autonomous navigation methods [40].

Laser Scanner: since the expected operation environment of Karo includes slopes, steps, stairs, and unstructured obstacles, the odometry method alone does not suffice for the robot's localization. Thus, for both localization and 2D-mapping of the system, a Hokuyo UTM30-LX LIDAR is prepared for the robot, which is mounted on a stabilizer to be parallel with the ground on inclined surfaces. As depicted in Fig. 13a, the laser scanner is connected to the robot platform's mini PC throughout the USB connection and in "Simultaneous localization and mapping" section we will discuss how this data is used for Karo's mapping.

Inertial Measurement Unit: we utilize the measurement regarding the changes in the attitude of the robot's platform using a 6-DOF inertial sensor, Xsens MTI-100, to control the laser scanner's stabilizer. The functionality of this stabilizer gets indispensable when the operation terrain is not even. We will discuss in "Evaluation of Karo" section regarding the highly uneven terrains on which Karo operates and accomplishes the exploration operations.

Thermal Camera: a thermal camera, Optris PI230, have been installed on the manipulator's end effector which is capable of synchronous capturing of visual and thermal images. This camera is used for detection and position estimation of victims during a mission as a part of the robot's exploration operations. It also ensures robot's visual acuity by providing thermographic information from targeted objects.

$CO_2$ sensor: we have prepared a gas sensor, MQ-9, which detects the presence of multiple types of gases in an environment. While this sensor has several applications for a rescue mission, we are more intended to verify a victim's breathing as a vital sign throughout this sensor. For instance, as illustrated in Fig. 23, Karo uses its sensor module including the gas sensor to explore the victim's hole.

Analog Cameras: Four analog cameras have been mounted on the robot's platform and its manipulator which provide a decent perspective for the robot's operator to monitor the robot, manipulator and surroundings by the installed cameras. Figure 16b demonstrates an image capture of the robot's graphical user interface (GUI) including the video streams of the analog cameras.

IP Camera: when the manipulator approaches a detected victim, the operator needs a high-resolution video of the location to complete the inspection task as clear as possible. To that end, a Sony high-resolution IP camera has been embodied into the manipulator's end effector and Fig. 16a illustrates how the IP camera provides a perceptive and broad sight of the car's ceiling.

Mono Microphone: two microphones have been installed on both the OCU and the manipulator's end effector to facilitate a full-duplex audio connection.

### Power systems

*Ac*cording to the schematic design of Karo's electrical systems, there are two separated power sources to supply the consumption of signal devices and actuators. The rationale behind this separation is the inherent difference in their consumption patterns. The signal devices, such as access points, cameras etc., have often a constant consumption through the whole operating time. On the other hand, the robot's actuators have significantly sporadic and dynamic consumption depending on the obstacles on the way or the operations in progress.

To provide the power source for Karo's devices and actuators, we first need to determine the systems specifications from consumption point of view. The power section of electronic devices needs a 24-V power source with 4 A continuous current. For the sake of the actuators, we similarly need a 24-V power source while its continuous current ranges from 0 to 20 A. The other factor to be considered is the system's burst current, which determines how quickly the batteries are going to be discharged. This factor is only depended on the actuators' behavior due to their stall current when they are applying their maximum torque. Accordingly, the c-rating of the power source must satisfy this characteristic of the system.

Considering the above discussion, we provided two 24 V lithium polymer battery packs with 10,000 mAh capacity and a c-rating of 10. Having said that, the power section of electronic devices can operate up to 2.5 h while the provided power source for actuator suffices for half an hour on average for each a mission. We will later on investigate the consumption pattern of the system and sufficiency of the provided power sources in "8.2" section.

### Hardware design
We need to design different ARM-based controller boards for both robot's platform and its manipulator to fulfil the schematic design of the control and command system, as illustrated in Fig. 13. All these controller boards benefit from LPC1768 Cortex-M3 microcontrollers with similar developed firmware. Since there are numerous specific requirements, constraints, and considerations regarding those controller boards, it was infeasible to choose a generic product in the market. Accordingly, we aimed to design those controller boards from scratch to satisfy Karo's system requirements as effective as possible. Here, we explain the functions of each developed hardware for both robot's platform and manipulator modules:

*The main controller:* the main controller board is installed on the mobile platform to (1) manage robot's power system, (2) communicate with motor drivers by sending the commands received from the OCU, and (3) control robot's peripheral equipment. To that end, various DC/DC convertors supply devices with different voltage levels such as 3.3, 5, 12, and 15 V. Besides, the board controls robot's front and rear lights, alarms, indicators, and the encloser's fans by a series of relays. To communicate with the motor drivers, the main controller relies on Controller Area Network (CAN) protocol, which provides relatively a robust communication between nodes on a mobile system. In addition, we implemented the hardware requirements for Ethernet protocol on the main controller board using a 10/100-Mbps Ethernet PHY to establish a reliable connection with the OCU.

*The manipulator's controller:* the manipulator control board mainly functions to control the manipulator's servo DC motors. To that end, it is equipped with the same Ethernet hardware as the main controller board to communicate with OCU. Further, it communicates with the servo motors through the RS-485 serial protocol.

*The end-effector board:* the end-effector board is installed inside the sensor box of the manipulator. Basically, it electrically handles all equipment and sensors in the sensor box such as controlling the LEDs, controlling the gripper's DC motor, data acquisition from gas sensor, supplying the speaker and cameras etc.

Motor Drivers: one of the most serious challenges in the hardware design of the control and command system is designing the motor drivers of the mobile platform's actuators. As elucidated in "Mechanical design" section, there are two Maxon RE 200 W DC motors for the robot's movement and two Maxon RE 150 W DC motors for the robot's flippers. Thus, the main challenge would be handling of both continuous and stall currents of these high-power actuators aligned with decent control and communication implementations in the motor drivers. Accordingly, we designed the motor drivers based on the H-bridge circuits for switching the motor's polarity via PWM signals. Practically, each of these custom-designed motor drivers supplies a load with 24 V and up to 20 A current, which confidently satisfies the system's requirements (Fig. 14).

### Operator Control Unit (OCU)
The portability, reliability, and connectivity of an OCU in a rescue mission significantly impact the robot's performance. By the way of the OCU, the operator should be able to control the movement of robot and manipulator, observe the environment with broad and wide perspective, monitor the robot's internal and external states, and interact with the robot's advanced functions such as mapping, exploration, object detection, etc. Accordingly, we designed and implemented the OCU embedded in waterproof and dust proof case, as shown in Fig. 15a, which can be set-up and break down the robot operation system in less than 10 min. In this regard, Fig. 15b delineates the operator remotely controlling Karo to climb up the stairs.

Besides, the OCU is facilitated by two packages of 10,000 mAh batteries that supply the whole system for more than an hour of operation. All other the OCU's devices and equipment have been depicted in Fig. 13c.

### Software systems
#### Human–robot interface
As the dexterity, mobility and exploration capabilities of the developed rescue robot presents higher complexity, operating the robot becomes more challenging and complicated. Accordingly, the way that the HRI is implemented can significantly affect the desirability of the robot's performance by providing an effective, convenient, and friendly interface for the operator. In the developed software system integrated with Robot Operating

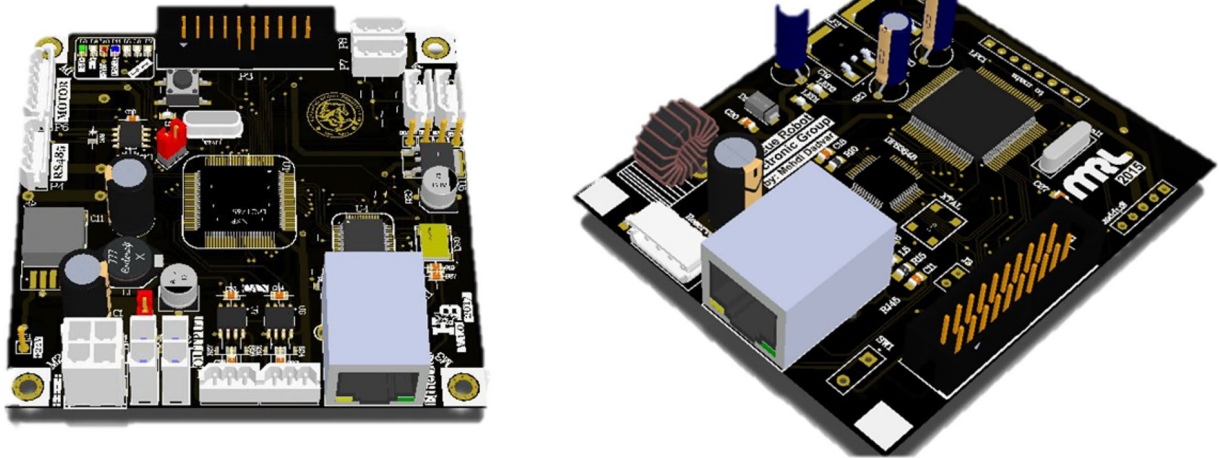

**Fig. 14** Hardware design of (left) the manipulator controller board, and (right) the end-effector board

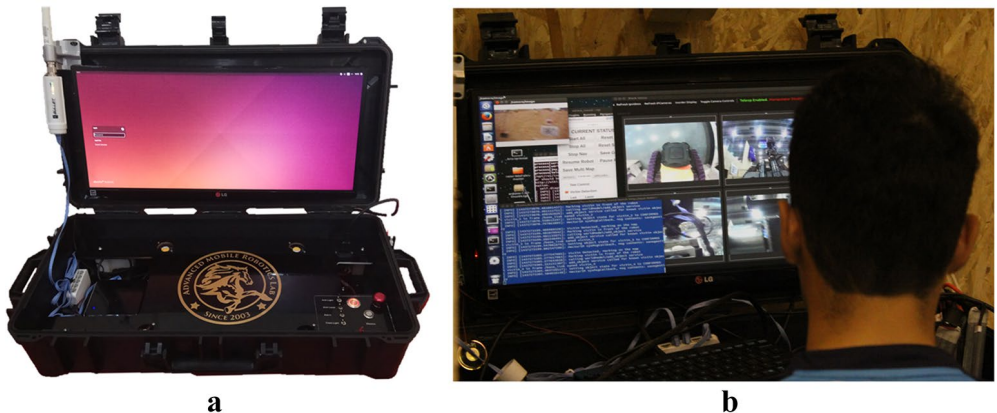

<p style="text-align:center">**a**</p>
<p style="text-align:center">**b**</p>

**Fig. 15** **a** The operator control unit of the robot, and **b** the operating controlling the robot

System (ROS) [41], we implemented the human supervision and control of the robots as two independent but closely packed software. Since the cores of these two software are different and it is inadmissible in ROS for a single node to connect to more than one ROS core, we tackled a software engineering challenge to create an integrated GUI for the whole system.

In the design of the HRI, we chose the colors of components in the robot's GUI in such a way that the color of each section naturally reflects its application. As a result, the operator can get use to the GUI's feedbacks and controls as effective and fast as possible. The GUI essentially serves for visualizing the sensory information regarding the state of the environment and the robot itself including video streams, gas sensor, thermal images, robot's power, movement, and joint states etc., as shown in Fig. 16. Moreover, the robot's GUI provides a dynamic and interactive visualization of the 2D map generated by the robot by employing the RVIZ plug-ins [42] of ROS. Besides, by utilizing ROS's RQT GUI plug-in, the interface's features can be evolved and customized as needed with no need to major changes in the underlying deployed codes.

The control commands for robot's mobile platform and manipulator are sent throughout an Xbox 360 wireless controller. The controller mainly serves in two mode (1) to control the robot's movement and position of flippers, and (2) to controller the robot's manipulator. The assignment of functions to buttons is a critical task to do since it must provide simultaneous control over multiple function of the robot for the operator as convenient and efficient as possible. Besides, we have provided a safety feature for the robot called armed/unarmed mode which prevents any unintentional commands from being sent to the robot. In addition to the Xbox controller, the commands can be sent using the keyboard or other controllers utilizing an abstract ROS node, which can be easily modified according to the type of the controller.

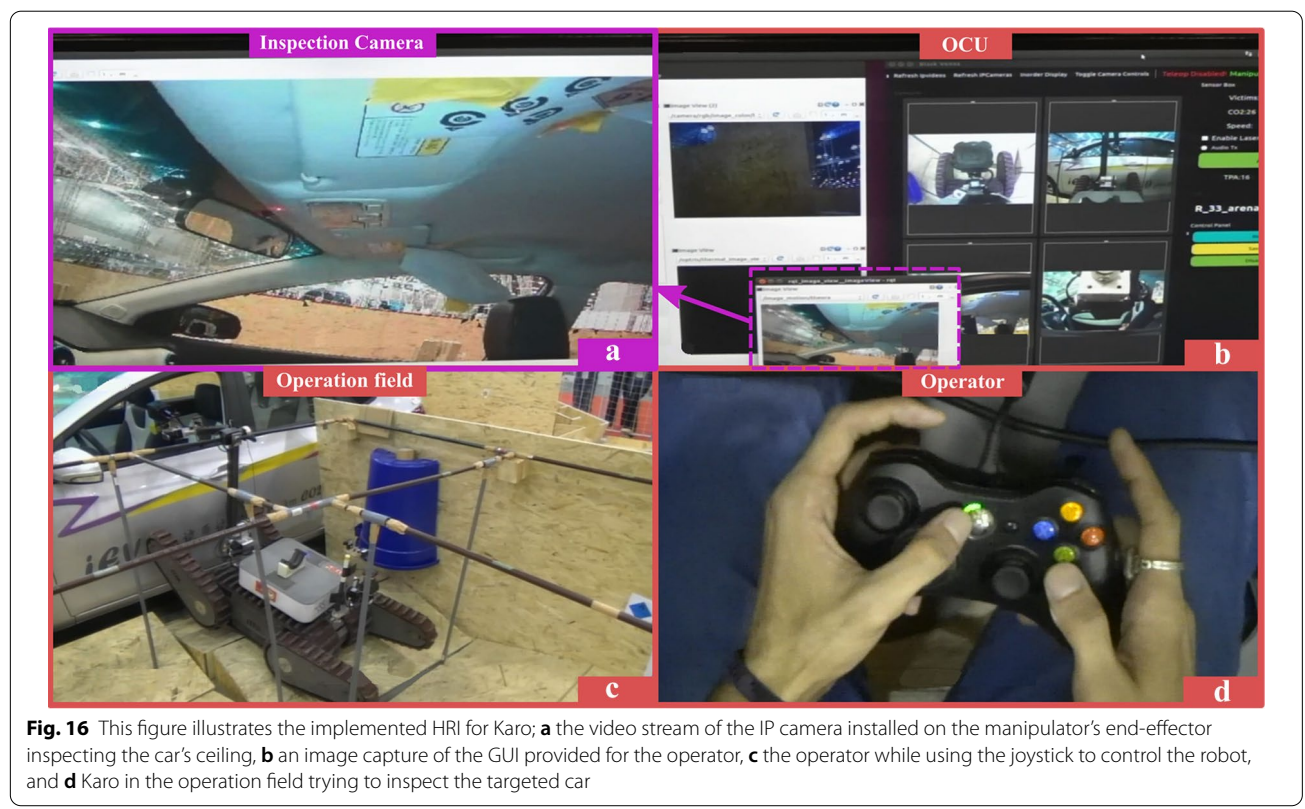

**Fig. 16** This figure illustrates the implemented HRI for Karo; **a** the video stream of the IP camera installed on the manipulator's end-effector inspecting the car's ceiling, **b** an image capture of the GUI provided for the operator, **c** the operator while using the joystick to control the robot, and **d** Karo in the operation field trying to inspect the targeted car

### Simultaneous localization and mapping
Rescue robots are expected to effectively explore the environment and accomplish the reconnaissance operations as well. Mapping is one the most imperative reconnaissance tasks assigned to rescue robots in a disastrous situation. Principally, generating a 2D map of the environment, in which the locations of victims have been uncovered, provides a decent perspective for rescue personnel to efficaciously progress the rescue mission. In this regard, the occupancy grid maps [43] have received adequate attentions throughout a wide variety of applications such as localization, path planning, and collision avoidance. Besides, there have been many works focusing on the SLAM topic [44], which is based on occupancy grid maps, as an important concept of mapping in mobile robotics.

To address the SLAM problem on mobile robots we need to deal with some serious challenges: (1) the USAR fields are covered by unstructured obstacles, stairs and steps, and inclined surfaces which makes normal planar indoor solutions [45] insufficient and inapplicable; and (2) most of the SLAM-based methods require accurate odometry data as a part of the input, while odometry data is inherently noisy and unreliable for localization purposes. Hence, we utilize a flexible and robust SLAM system with 6-DOF motion estimation [46] for the robot's 2D map generation. Furthermore, a precise LIDAR system with high update rate, as explained in "5.2" section, has been installed on the robot to enhance the accuracy of the mapping. To cope with the disturbance of uneven terrains, a 2D stabilizer compensates the changes in the attitude of the laser scanner and robot's platform, caused by the slope gradient, throughout the read data from the IMU. In this regard, Fig. 17 illustrates a 2D map generated by the robot in an extremely rough environment. As a matter of fact, this map is a merged 2D map by Karo and another complementary robot in the mission. Map merging capability of Karo gives an overall and decent perspective of the environment to the rescue personnel which is more effective and useful than giving them multiple maps with partial observations. This has been done

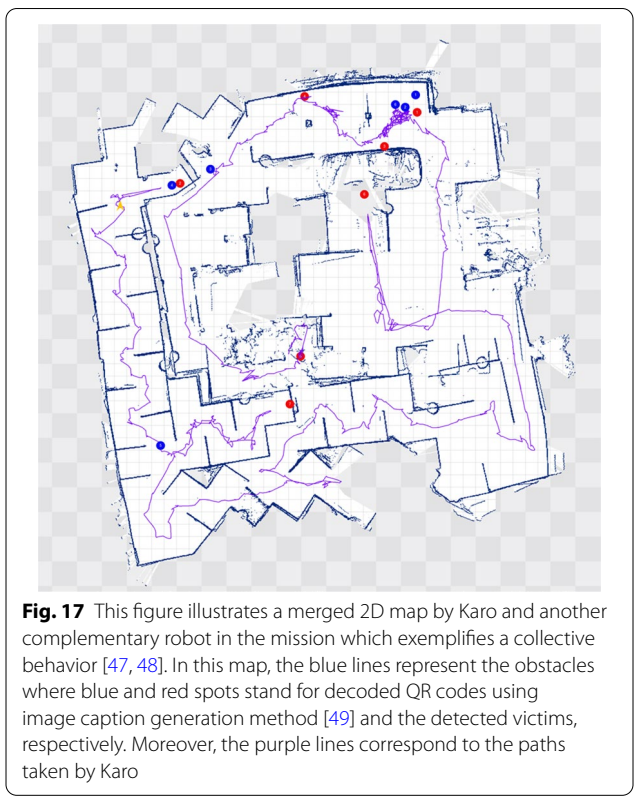

**Fig. 17** This figure illustrates a merged 2D map by Karo and another complementary robot in the mission which exemplifies a collective behavior [47, 48]. In this map, the blue lines represent the obstacles where blue and red spots stand for decoded QR codes using image caption generation method [49] and the detected victims, respectively. Moreover, the purple lines correspond to the paths taken by Karo

by finding transformation between two maps, generated by two different robots, and merging them accordingly.

#### Autonomous exploration

*To* take a step towards autonomous behavior of the robot, we implemented a simplified autonomous exploration utilizing the existing sensors and devices on the robot. By a "simplified" implementation we mean that the control of flippers still needs to be tele-operative and the algorithm controls only the mobile platforms movement. Having said that, this function does not suffice for performing autonomous exploration in excessive uneven terrains such as environments with stairs, ramps etc. However, the autonomous exploration function of the robot can enhance the robot's performance in flat corridors or terrains with moderate obstacles such as the "Crossover" test of the RRL's maneuvering test suites or "Map on Continuous Ramps" from exploration test suites.

The robot in the autonomous mode is expected to explore the unknow environment and perceive the surroundings for detecting victims, QR codes, specific objects, and other points of interest. To that end, we developed a highly modular system which benefits from a multi-layer architecture including behavior control, global path planner and trajectory generation layers. Here, we briefly describe each of these layers:

Behavior control: behavior control functions as a high-level controller and determines the robot's objective, which is approaching a potential victim or exploring the environment.

Global path planner: the global path planner generates a path regarding the goal set by behavior control layer in order to maximize the coverage over the environment and minimize the distance travelled.

Trajectory generation: when goal point has been determined and a path is generated for reaching that goal, the trajectory planner generates trajectories that can guide the robot to follow the generated path till it gets to the goal point considering the current states of the robot.

In practice, there are many complexities that makes the explained implementation challenging. For instance, the environment is initially unknown for the robot. Thus, it is infeasible for the global path planner to find the best path in a partially observable environment. However, as the robot keeps exploring utilizing the frontier-based method [50], the environment becomes more observable for the robot and the output of the global path planer converges to an optimal solution. In despite of the abstractions and simplifications of the autonomous exploration function implemented for the robot, its performance turned out to be an effective complement for the robot's tele-operative mode especially in flat and wide corridors. For instance, it is assertively feasible to employ this autonomous function in two of the Maneuvering test suites to enhance the robot's overall performance, as will be discussed in "Evaluation of Karo" section.

#### System specifications

Basically, this section mediates between the presented design procedure of Karo and the experimental results. In other words, the robot's specifications presented in this section stand for the theoretical expectations which has to be practically investigated by a wide variety of experiments and field evaluations. Table 3 illustrates the robot's specifications mainly focused on robot's movement, power system, manipulation, and application. In fact, these specifications are the building blocks of the robot's performance from mobility, dexterity, and exploration points of view. For instance, the set of sensors that the robot is equipped with shapes its exploration capabilities, which will be discussed in the next section. However, these sets of specifications provide a comprehensive

**Table 3 Karo's Specifications**

| Robot's specification | Value |
|---|---|
| Name | Karo |
| Typical operation size | 0.8 * 0.6 * 0.6 m |
| Transportation size | 0.9 * 0.8 * 0.7 m |
| System Weight | 85 kg |
| Weight including transportation case | 100 kg |
| Unpack and assembly time | 210 min |
| Locomotion | Tracked |
| Assistive mechanisms | 4 flippers |
| Maximum speed (flat/ outdoor/ rubble pile) | 0.8 / 0.6 / 0.3 m/s |
| Turning Diameter | Zero |
| Payload (flat) | 30 kg |
| Power source | Lithium Polymer batteries |
| Battery Endurance (idle/ normal/ heavy load) | 90 / 40 / 20 min |
| Batteries' charge time (80%/ 100%) | 45 / 60 min |
| Manipulator | 2 link arm, DOF 7 |
| Manipulator's reach (vertical/ horizontal) | 130 / 130 cm |
| Manipulator's payload at full extend | 5 kg |
| Sensors | four analog cameras, one high resolution IP camera, thermal camera, gas sensor, IMU, Laser Scanner, and microphone |
| Communication | IEEE 802.11a 5 GHz 500mW |
| Tether | optional |
| Operation modes | Tele-operative and semi-autonomous (flat ground) |

comparison with robots investigated in DHS/NIST sponsored evaluation exercises [30].

## Experiments

### Locomotion mechanism experimental results

In order to confirm that the purposed analysis of "Locomotion mechanism" section assisted motor selection, we tested Karo while performing both scenarios and measured the drawn current from each motor during motion (Fig. 18). Particularly, these tests verify our prior considerations for power loss due to the undetermined frictions. It should be mentioned that the robot tested with full equipment, the manipulator is not shown in Fig. 18. Since each of DC motors has a unique mechanical specification (i.e., torque constant and nominal current), the corresponding load torque at each measured current value is calculated by using Eq. (11), taken from the Maxon Motor datasheet, to verify that both motors operate within the suggested continuous range by the manufacturer company. The torque constant $K_m$ is 38.5 mNm/A and 30.2 mNm/A, and the nominal current $I_0$ is 236 mA and 137 mA for the 200 W and 150 W DC motors, respectively.

$$I = \frac{M}{K_m} + I_0 \qquad (11)$$

Figure 18 illustrates the succession of Karo in climbing a 40° ramp and drawn current/torque from the motor in each sequence. Sequences 1 and 4 both include a period that the robot does not have any motion (0−1 s and 9−10 s), and a period of moving on a flat surface (1−3 s and 7−9 s). Measured current values during the motion periods are about 2 A, which is the indication of existing friction between the robot and the surface. As the robot reaches the ramp (Sequence 2), the current increases until it is completely on the ramp (Sequence 3), which draws the maximum current from the motor. For reasons unknown, there are irregular spikes in measurements where the robot's body enters the inclined surface and leaves the bottom landing. However, there are several possibilities: the slippage between tracks and the floor; uneven wear on tracks; and possible shocks during the transition between surfaces. Although some of the measurements pass the 10.8 A (405 mNm) maximum continuous current (torque) of the 200 W motor, the overall performance

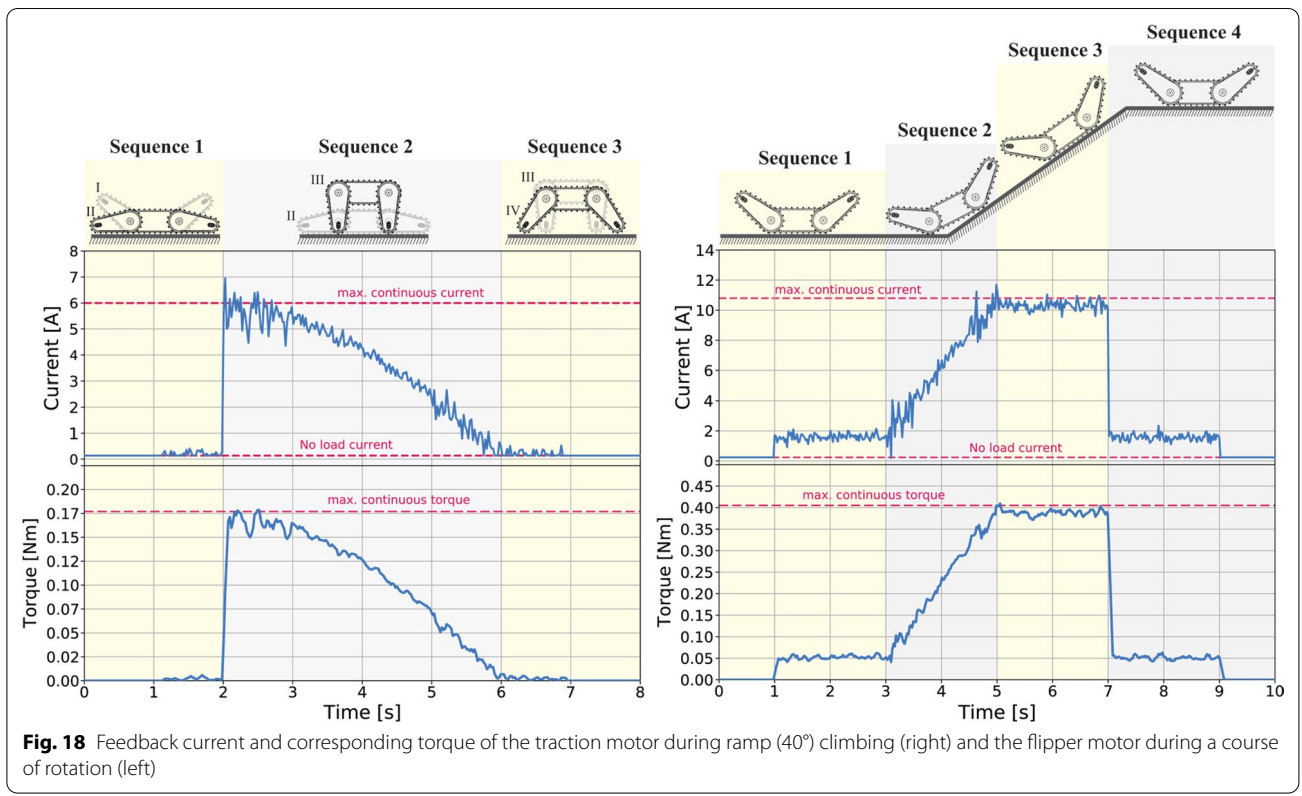

**Fig. 18** Feedback current and corresponding torque of the traction motor during ramp (40°) climbing (right) and the flipper motor during a course of rotation (left)

does not exceed the continuous operation range, so the robot can smoothly operate in various conditions. This can be seen clearly from the 8 A difference between the measured current on a flat surface and an inclined one. Additionally, climbing ramps or stairs at (40°) angle happens for a short period and it is considerably less frequent during operation.

Similarly, Fig. 18 depicts Karo's flippers lifting and putting down the robot in three sequences, starting at $\beta = -45°$ and finishing at $\beta = 45°$. The lower Greek number indicates the starting position, and the higher Greek number indicates the finishing position in each succession. There is no movement in the first and last seconds of motion. In Sequences 1 and 3, the robot/flippers' weights help the motors to rotate effortlessly, so the demanding current or torque from the system is very low. The unknown frictions impose less than 1A to each motor. The highest current was measured when the flippers contact the floor at the beginning of Sequence 2. There are similar unidentified spikes in this transition, which may be caused by the large contact surface or slippage between tracks and the floor. It is important to know that the peak current happens in less than 1 s every time the flippers lift the robot. This indicates that flippers are certainly capable of assisting the motion in various conditions in which the robot overcoming obstacles or trapped between objects.

## Analysis of batteries' discharge

The power system of remotely operated robots plays an imperative role in their functionality and effectiveness while completing an assigned mission, especially when they are equipped with on-system power sources. Correspondingly, we need to validate the adequacy of Karo's power system by investigating its durability in various standard missions (Center, Crossover, Travers, and Stair Debris). We selected these four different RRL's tests as the testbench for measuring the batteries discharge pattern. We know that Karo would have the highest consumption in Stair Debris test compared to other selected tests, since its movement actuators require higher torques to overcome the obstacle. Pragmatically, we seek the durability of Karo consistent with the RRL's 30-min missions for the worst-case scenario of the robot's consumption (Stair Debris test). As the robot is equipped with two battery packs for its actuators and electronic devices, we conduct the experiment for both power resources separately.

Figure 18a illustrates the discharge measurements for the actuators' battery pack in 60 min. As we expected, the batteries discharge in Stair Debris test earlier compared to other tests, though they endure 42 min (i.e., 12 min more than required durability for RRL's missions) and satisfy the primary requirements. Besides, since the Center test only includes flat terrain, the battery discharge rate is relatively slower which ensures more than one hour (around 74 min) of continuous operation for the robot. We observe that the durability of the robot's batteries decreases by increasing the obstacles' complexity in the selected tests, where the batteries' durability dropped 43% from Center test to Stair Debris test.

The same measurement has been applied to the battery pack of the electronic devices, as shown in Fig. 18b. In despite of robot's actuators, the electronic devices have almost a constant power consumption regardless of the type of the mission. As a result, in all four tests the battery's discharge patterns are significantly analogous and demonstrate a drop of voltage from 12.6 V (fully charged battery's voltage) to 11.45 V after 60 min of operation, where the discharge voltage is 9 V (Fig. 19).

### Karo's performance in RRL
#### Evaluation of Karo
In this section we aim to evaluate the Karo comprehensively in terms of maneuvering, mobility, dexterity, and exploration capabilities. As we stated in "Introduction" section, the main motivation of this work is to embed all four complementary capabilities in a single rescue robot for accomplishing an effectual USAR mission. In this regard, the RRL test suites inspired from DHS-NIST-ASTM international standard test methods facilitates this aim by offering methods to evaluate rescue robots quantitively. Accordingly, we tested Karo in all four test suites meticulously based on the rules and frameworks of RRL. To draw a conclusion about the results confidently, we repeated each test 20 times where each test was conducted in a 30-min time slot. Thus, we will discuss the results and their variation for each test first, and then we will compare the Karo's performance with other robots taken part in RRL.

Maneuvering is the most basic capability required for accomplishment rescue missions. Figure 24a illustrates the results of 20 tests for each Maneuvering test taken by the Karo. Although the Maneuvering tests have no major mobility challenge for the Karo, as shown in Fig. 20 for the Traverse test, maneuvering in confined corridors is an exacting task to accomplish for maxi-sized robots. Since the RRL's rule allow five more minutes only for autonomous operations, we utilized the autonomous exploration capability of the robot in Center, Continuous Ramps, and Crossover ramps of the Maneuvering test suite. As a result, the robot completed up to two more repetitions in those tests.

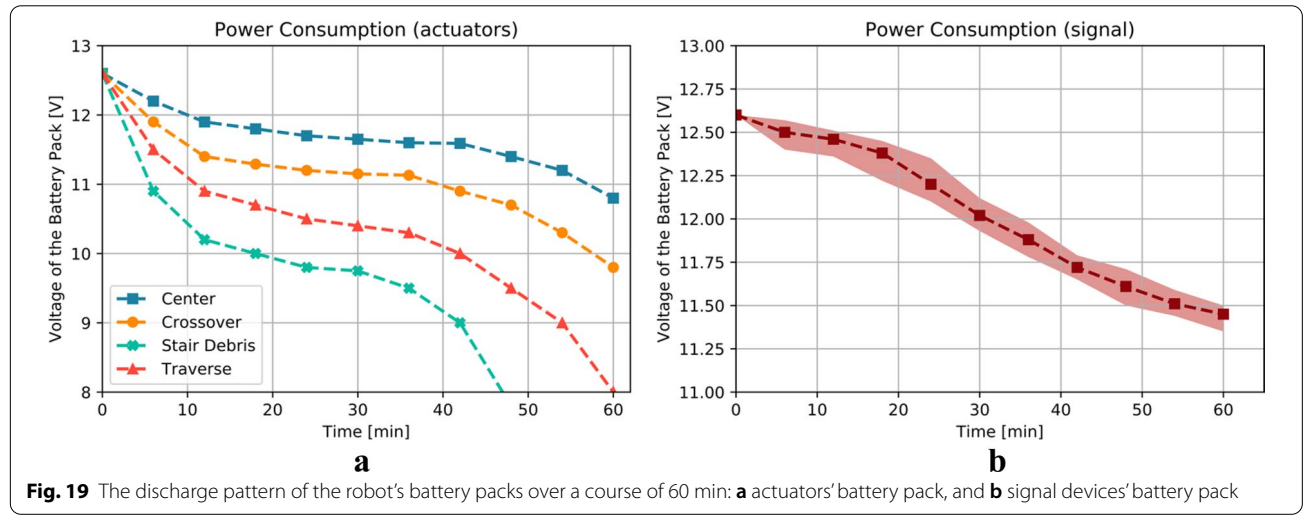

**Fig. 19** The discharge pattern of the robot's battery packs over a course of 60 min: **a** actuators' battery pack, and **b** signal devices' battery pack

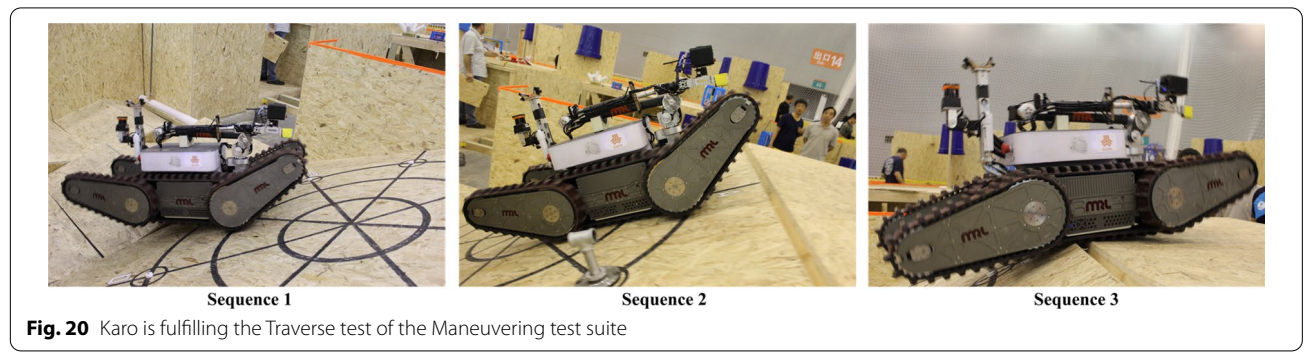

**Fig. 20** Karo is fulfilling the Traverse test of the Maneuvering test suite

The mobility capabilities of Karo have been evaluated in five Mobility tests based on the RRL's framework, as demonstrated in Fig. 24b. Statistically speaking, the robot performed the best on the Hurdles and Stair Debris, which is one of the most challenging mobility tests. The results of the mobility tests imply that the robot benefits from an immense stability while overcoming difficult obstacles, although its performance on Sand/Gravel Hills test is ordinary. Moreover, Karo also accomplished a satisfying operation on the Stepfields test, as shown in Fig. 21.

Similarly, the dexterity skills of Karo have been evaluated quantitatively throughout five Dexterity tests. As illustrated in Fig. 24c, Karo has accomplished all Dexterity tests fulfilling with small variations in the 20 repetitions for each test. One explanation can be the similarity of the test elements in parallel pip, omni-directional, and cylindrical pipes tests. In this regard, Fig. 22 demonstrate Karo's performance while inspecting inside parallel pipes. Beyond the RRL's dexterity test suit, Karo's dexterity capabilities have also been evaluated in a complementary testbench, where the robot must rely on both its mobility and dexterity skills to inspect holes in a highly uneven environment. As shown in Fig. 23, Karo is inspecting a hole on uneven Stepfields to detect a simulated victim.

In contrast to all other evaluations, Karo's exploration capabilities have been examined via only three exploration tests. That is mainly because two tests of Exploration test suite (i.e., Avoid Holes and Avoid Terrain tests) are designated to fully autonomous robots. Since Karo's autonomous exploration function is not adequate for those two tests, its exploration skills have been tested only for Map on Continuous Ramps, Map on Crossing Ramps, and Recognize Objects tests, where Fig. 24d depicts the results. Considering the results, Karo's performances on the first two exploration tests are highly identical. To elucidate, provided terrains in both tests are sort of unproblematic for Karo's mobility skills and as a result, practically speaking, there is no noticeable difference in these two tests.

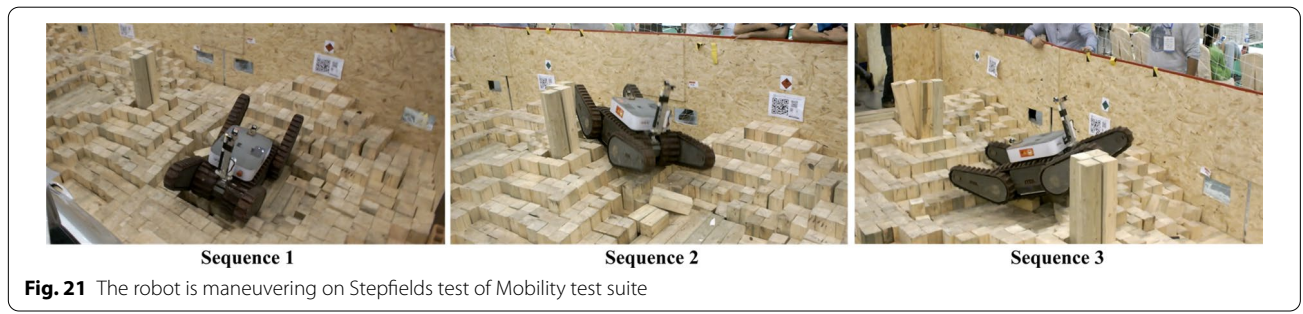

**Fig. 21** The robot is maneuvering on Stepfields test of Mobility test suite

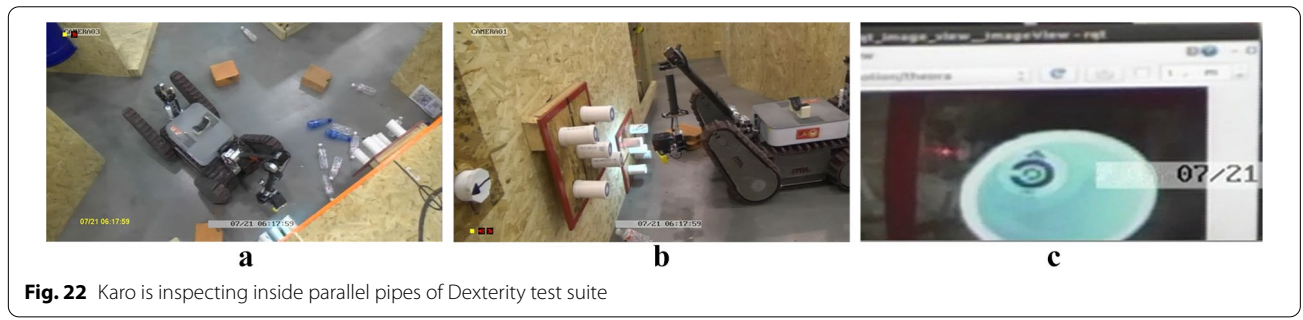

**Fig. 22** Karo is inspecting inside parallel pipes of Dexterity test suite

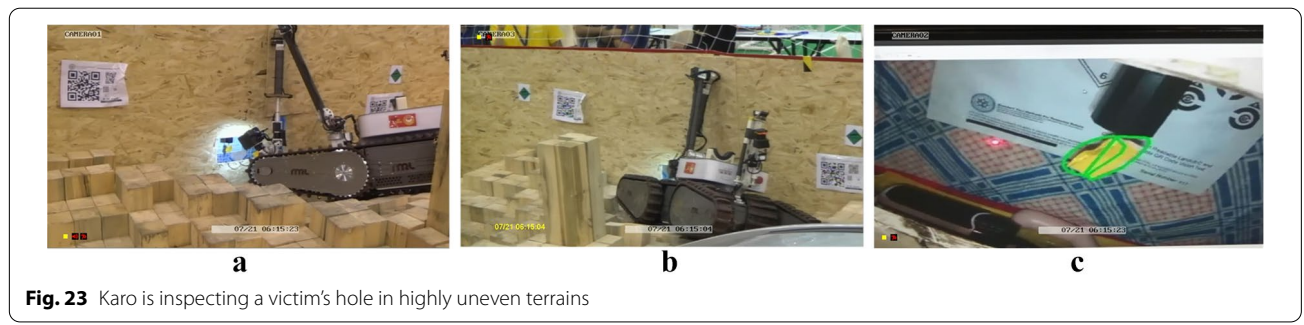

**Fig. 23** Karo is inspecting a victim's hole in highly uneven terrains

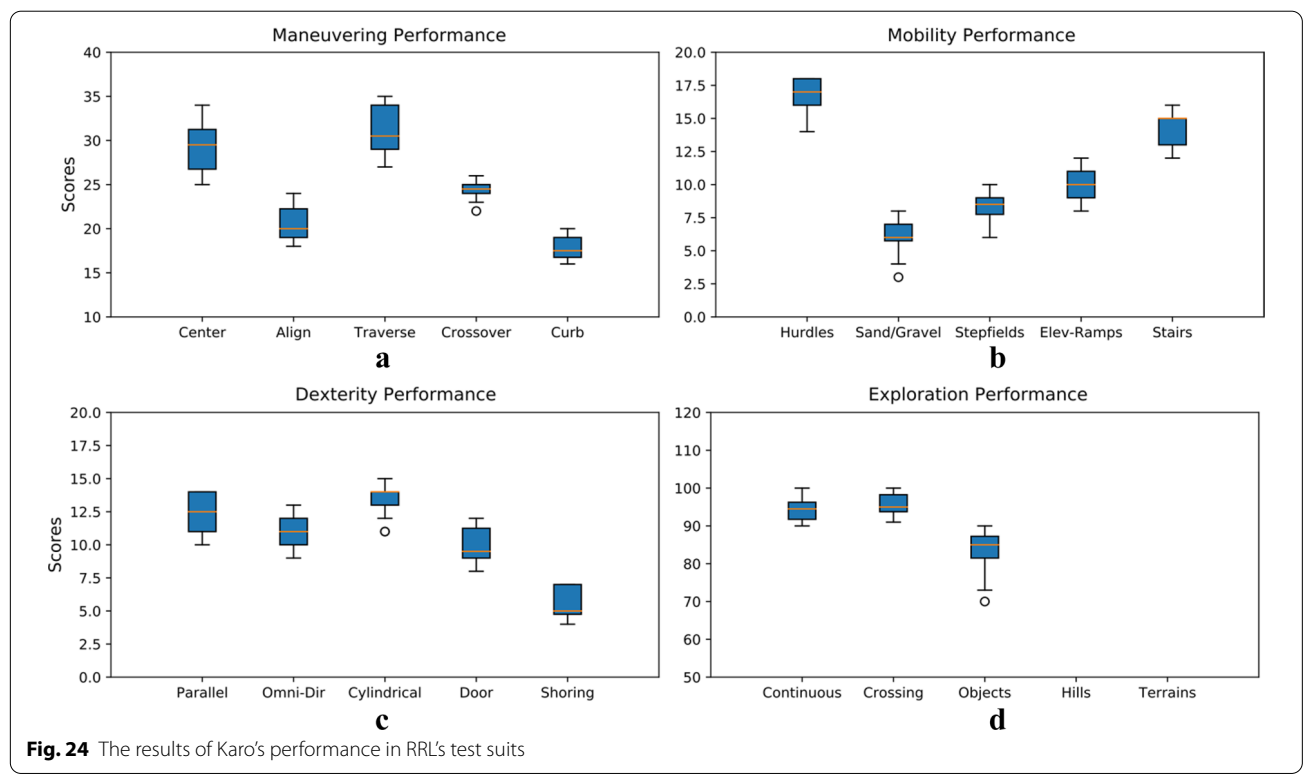

**Fig. 24** The results of Karo's performance in RRL's test suits

***Comprehensiveness of Karo's Capabilities***

In this section we aim to compare the performance of Karo with other robots participating in RRL. We will draw the comparison from two point of views: 1) comparing Karo with the average performance of participating teams in RRL, and 2) comparing Karo's comprehensive performance with robots demonstrating the best performance in each test suite. The former provides insights about Karo's performance in each individual test and compares its capabilities with the average results of all other teams participated in RRL based on the same test frameworks. The latter examines the comprehensiveness of Karo's capabilities with respect to the best performances in RLL in each test suite. Altogether, not only we ensure that Karo is functioning significantly superior to the RRL's mean performances, but also, we demonstrate that all required capabilities have been embedded in Karo adequately with respect to the best RRL results.

Figure 25 illustrate the comparison from the first point of view, in which Karo have done far superior to the mean performance of all robots in RRL. The only exception is in the exploration test suits where Karo has not participated in two tests since they are designated to fully autonomous operations. However, the gap is sizable in the rest of the exploration tests. In addition, the RRL's mean performance is closer to Karo's performance in the Maneuvering test suite than in any other test suite. That is mainly because the Maneuvering test suite is the most rudimentary evaluation of rescue robots competing in RRL and all robots must take all Maneuvering tests to demonstrate their basic capabilities. As a matter of fact, the results in Fig. 25 have not been normalized and represent the raw score collected by robots.

In Fig. 26, a comparison is drawn from the second point of view, as explained above. In this comparison Karo has been compared with three other rescue robots, each winner of one class such as Mobility, Dexterity and Exploration. In this comparison, we are considering normalized scores of all five tests in a test suite in order to investigate the robots' comprehensive capabilities. In all three comparisons, Karo has done comprehensively more satisfying performance compared to its opponents. For instance, concerning Fig. 26c, although the winner of the Exploration has accomplished a cut above Karo in one test suite (i.e., exploration), in all other three test suites Karo has demonstrated higher quality operations. Numerically speaking, the same discussion can be drawn for Fig. 26 a, b.

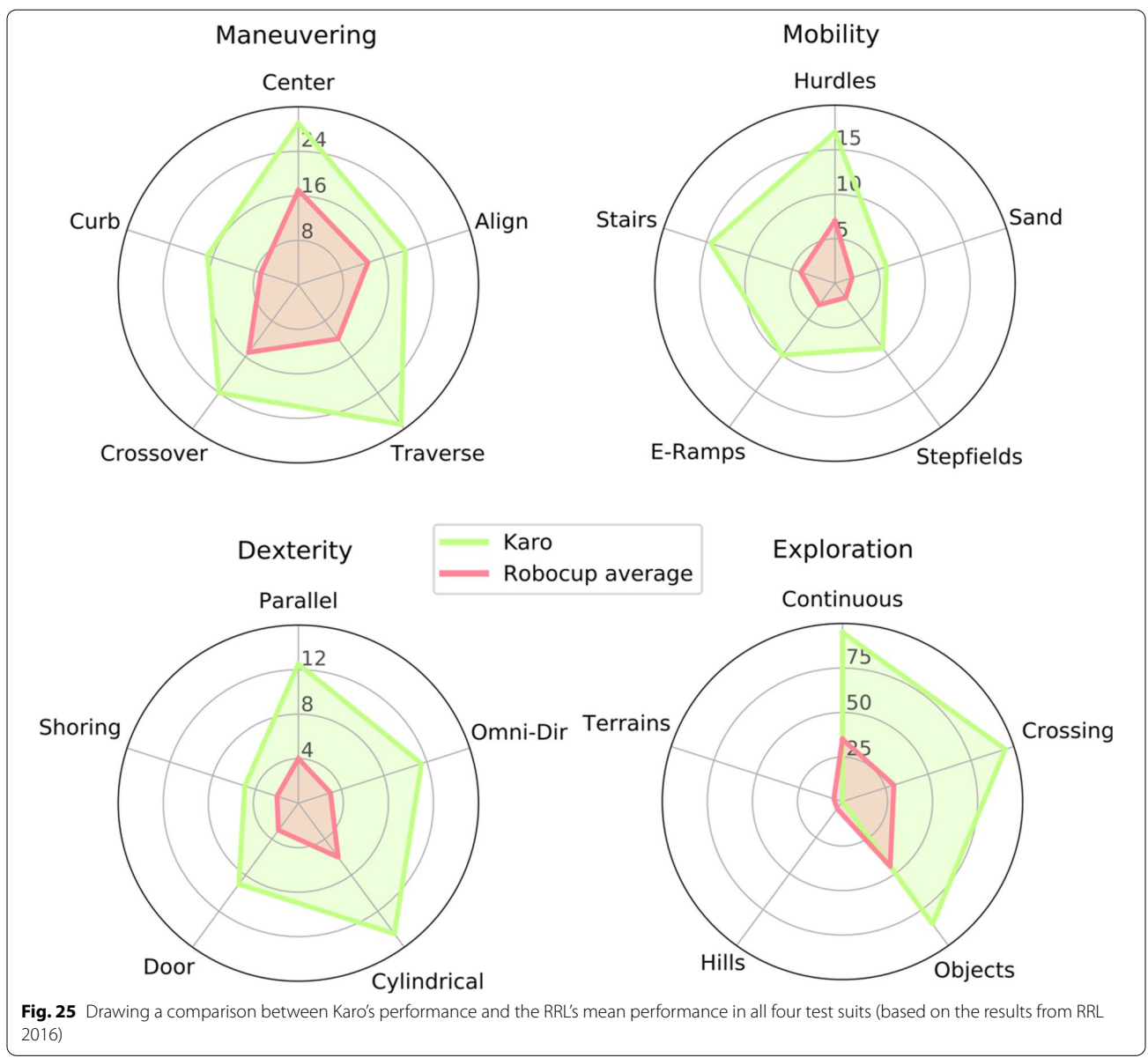

**Fig. 25** Drawing a comparison between Karo's performance and the RRL's mean performance in all four test suits (based on the results from RRL 2016)

**Field performance**

Motivated to bridge the evaluation gaps of RRL's framework inspired by NIST standard test methods, this section discusses the test and evaluation of Karo at the training field of a fire department as its field performance. Generally speaking, standardized test methods including RRL put forward structured and reproducible test suites that cannot reflect the haphazard pattern of obstacles and corridors in the real world's disaster sites. Besides, the design of participating robots can be biased towards the RRL's test suites. As a consequence, the performance of those robots remains uninvestigated in unprecedented and less frequent scenarios in the real world's problems that has not been modeled by any of those standard test suites. By the same token, the performance of a robot's operator can be biased to the foreseeable structure of the RRL's test suites, while the operator has to make unplanned and spontaneous decisions in unpredicted scenarios in more realistic situations. Considering these evaluation gaps, the field performance tests have been conducted to validate Karo's performance in mobility, inspection, endurance, and communication in a more generic and unrehearsed manner. These tests challenging the robot's practical capabilities help the developers to (1) challenge the robot's capabilities beyond the scope of RRL, and (2) gain insights into the future designs and developments. Although the robot failed in three validation tests as illustrated in Table 4, its promising comprehensive performance adequately satisfied the mission's objectives. Here, we present a detailed elaboration of each test item conducted in 30-min time slots.

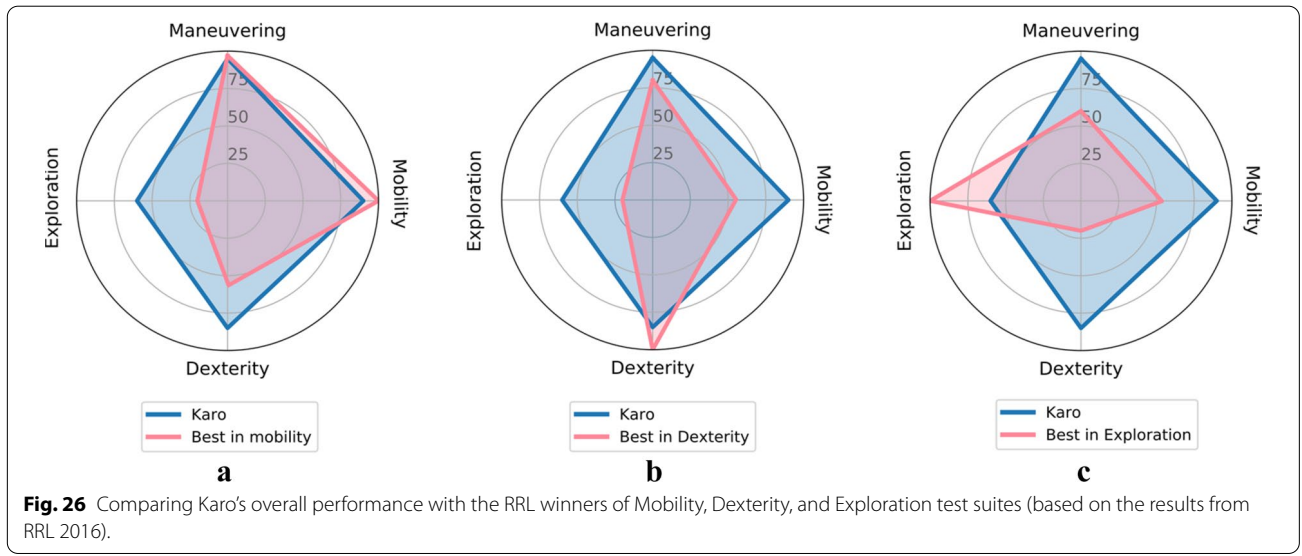

**Fig. 26** Comparing Karo's overall performance with the RRL winners of Mobility, Dexterity, and Exploration test suites (based on the results from RRL 2016).

In mobility tests, each test item is considered completed/ passed when the robot successfully fulfills two repetitions over the test terrain. Having said that, the robot effectively performed multiple mobility tests such as maneuvering on scaffolds, climbing up and down stairs with debris, traversing ramps and overcoming unstructured obstacles. Among all these mobility tests, stair with debris is the most challenging one which questions the robot's mobility and stability and the operator's adeptness instantaneously. As Fig. 27 demonstrates, the robot successfully completed multiple trails on the stairs with debris where the width and slope of the stairs are almost similar to the RRL's test suite. However, the non-wooden material of the structure and haphazard distribution of debris along the way throws up relatively more serious mobility challenges compared to the RRL's stair test suite. According to the results, the stair test with harsh debris aborted due to crossing the motor drivers' current limit. As a final note on the robot's mobility, its performance on different test items designed originally for human first responders was satisfactory, though the robot's depreciation was significantly higher compared to performing on the RRL's test suites.

**Table 4** Validation tests taken by the robot in the training field

| Validation test | Short description | Passed/failed |
| --- | --- | --- |
| Mobility | Maneuvering over a flat surface structured by scaffolds | Passed |
| | Climbing up and down the stairs with harsh debris | Failed |
| | Climbing up and down the stairs with moderate debris | Passed |
| | Traversing ramps (< 45°) | Passed |
| | Overcoming unstructured obstacles | Passed |
| Inspection | Inspecting inside targeted cars | Passed |
| | Inspecting areas below the ground level | Passed |
| | Performing inspection in dark corridors | Passed |
| | Performing inspection in smoky corridors | Failed |
| Communication | Communicating in line-of-sight situation (> 800 m) | Passed |
| | Communicating in non-line-of-sight situation (> 100 m) | Passed |
| | Communicating in building through different stories | Failed |
| Endurance | Operating continuously (> 1 h) | Passed |

Next, we put the inspection capabilities of the robot to the test throughout four practical tests namely car inspection, dark corridor exploration, smoky corridor exploration, and inception of areas below the ground level. An inspection test is considered completed when the robot approaches the determined target closely and then transmits acute and perspicacious information of the target. In this regard, the robot successfully inspected the space inside the car from distance enjoying the manipulator's ample workspace (see Fig. 11), as illustrated in Fig. 28.

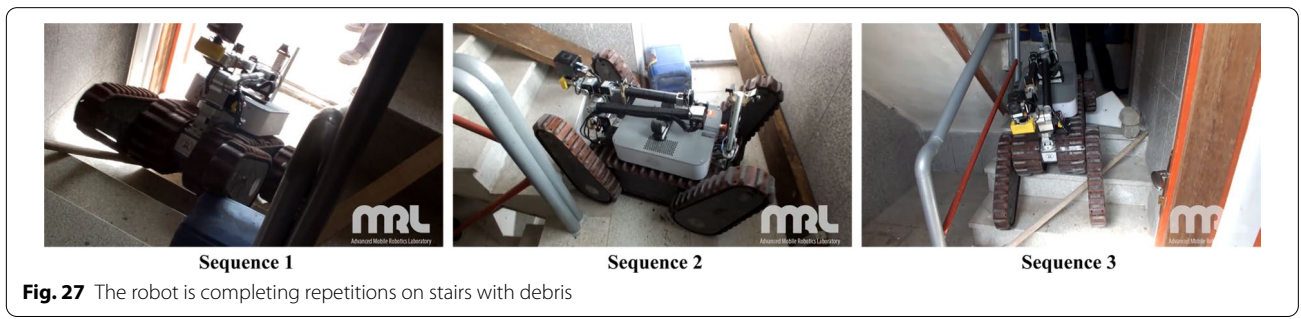

**Fig. 27** The robot is completing repetitions on stairs with debris

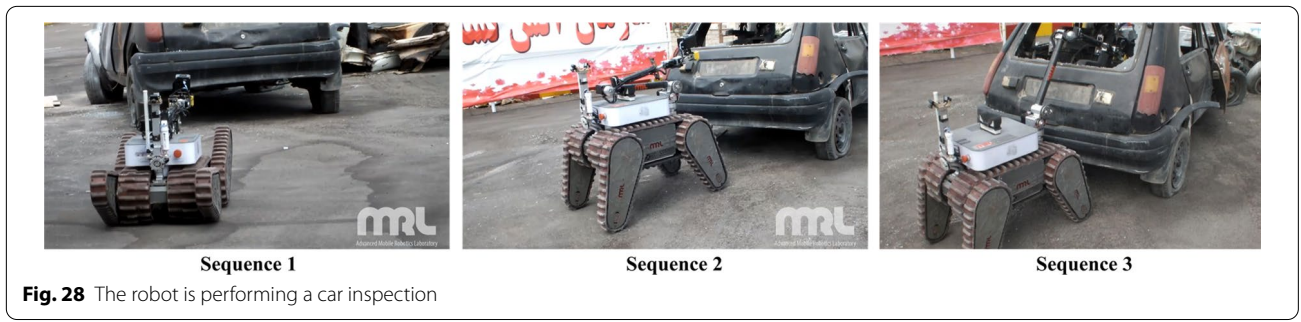

**Fig. 28** The robot is performing a car inspection

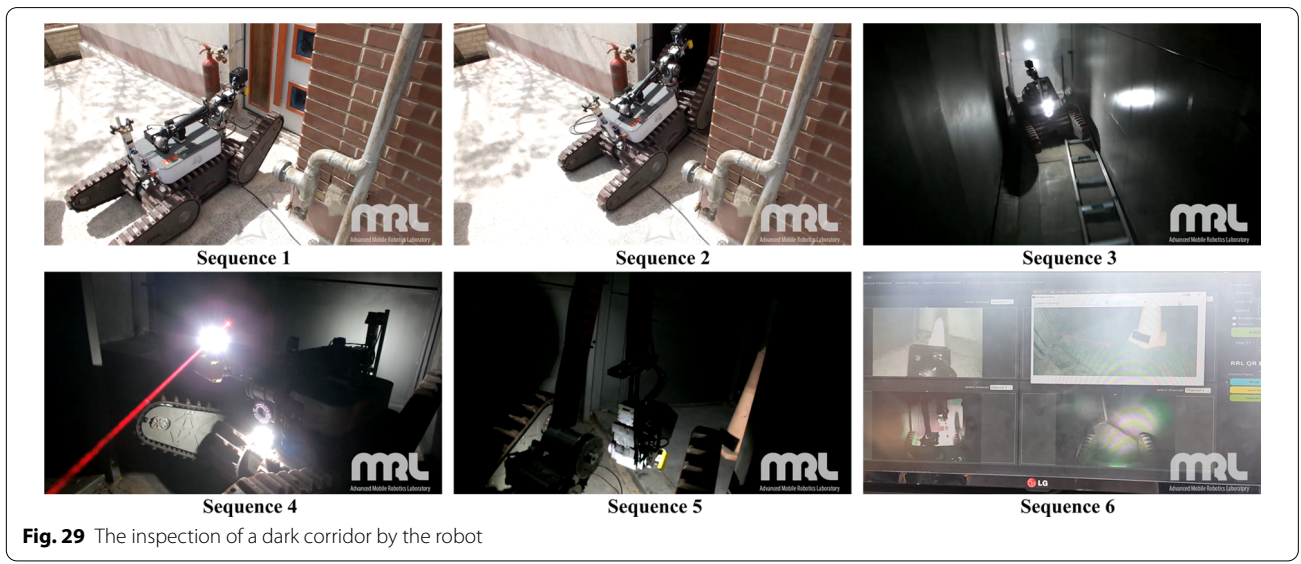

**Fig. 29** The inspection of a dark corridor by the robot

Regarding the corridor exploration test, the robot effectually explored the dark corridor utilizing its night-vision equipment, as shown in Fig. 29. However, it failed to maneuver in the smoky corridor because of the high density of the smoke which made the corridor almost unobservable for the robot. Relying on the laser scanner's multiecho function, we tried to utilize the online map obtained by the robot's laser scanner to navigate the robot through the corridor which failed similarly.

Subsequently, the inspection test displayed in Fig. 30 exemplifies scenarios in which the manipulator must have sufficient reachability to inspect the areas beneath its platform. Karo demonstrated its dexterity capabilities by accomplishing the test successfully. This capability of Karo's manipulator has been completely analyzed in "Achievable workspace" section.

During all mobility and inspection tests, the quality of wireless communication between the robot and the OCU has been constantly monitored to corroborate its functionality. The communication maintained reliable in line-of-sight situation up to 900 m. In the case of non-line-of-sight situation, the communication maintained robust for a 120 m-distance. However, the reliability of the communication was not ensured when robot was operating on the third floor of the building and the OCU was on the first floor. Although the OCU and the robot were still connected, the delay in the control commands had increased the risk of fault and failure of the operation.

A screen capture video of Karo's field performance can be found as a supplementary material along with this paper, by using the YouTube link "y2u.be/UnbqkQWR7e4," or by scanning the following QR code.

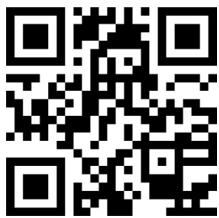

## Conclusion
This paper represents the methodology to design and implement a maxi-sized rescue robot (Karo) that is capable of carrying out the desired capabilities of response robots in real-world missions: mobility, dexterity, and exploration. Our approach is based on RRL's evaluation benchmark, obtained from the DHS-NIST-ASTM international standard test methods, to determine the system requirements. We comparatively analyze the conceptual design of our previous robots and therefore introduce an improved platform for Karo. We employ a streamlined approach to implement Karo's conceptual design that enables reliable operation during USAR missions. The experimental results and test evaluations of our work confirm that incorporating the knowledge about the real-world necessities into the design and implementation process enables tailoring a comprehensive response robot. We also highlighted that the important difficulties associated with real-life missions are not fully addressed in the standard tests, but they are significantly beneficial for response robot developers to begin with. We envision that the presented approach lays a foundation method for developers to strengthen the future generation of rescue robots.

To identify Karo's mechanical characteristics, we first developed a conceptual design for the locomotion system (tracked platform with four flippers) conjunct with a suitable joint arrangement (7-DOF) for the manipulator. Due to the unpredictable circumstances of rescue missions, we identified the operation range for motors accordingly and considered the extreme payload condition as a baseline for the continuous operation of our system. Our approach is verified experimentally, through comprehensive evaluation tests in RRL. Here, we employed a diverse set of sensors and perception devices for vision, audio, localization [51], and mapping, to control the robot from a remote operator control unit. Furthermore, we implemented a ROS-based software system for the GUI to enable the operator to monitor live sensory information in an environment besides providing real-time control of the robot (Fig. 16).

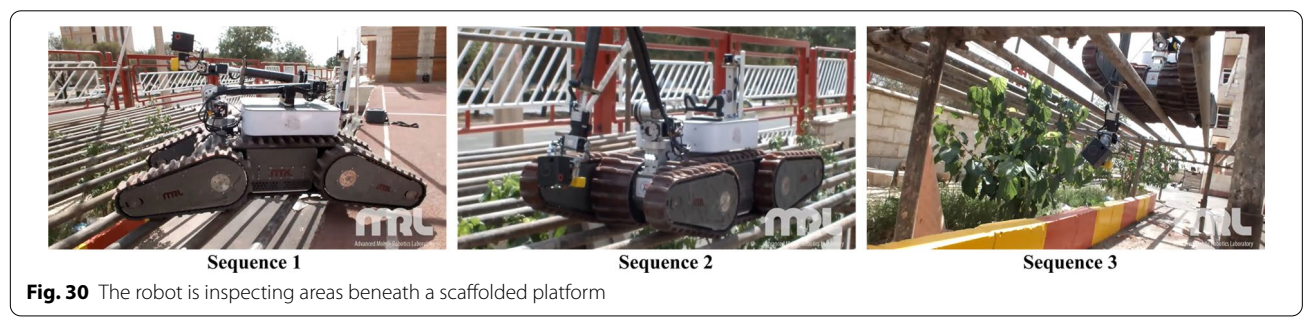

**Fig. 30** The robot is inspecting areas beneath a scaffolded platform

In this paper, through various standard test evaluations, we demonstrated that Karo's overall performance is not only improved substantially compared to its counterparts but also is superior to the average performance of RRL's other participants. Furthermore, RRL evaluations verify Karo's extensive capacity in all test suits, however, the top participants in RRL only performed well in some of the test suits. This highlights the importance of using the benchmark since a real rescue mission is indeed a combination of mobility, dexterity, and exploration. Lacking any of those capabilities could affect robots' performance drastically during missions. For instance, a robot with inadequate dexterity cannot complete a mission that requires object manipulation such as opening a door. Therefore, various capabilities should be considered in the development of a response robot.

As a final demonstration, we validated our approach more practically through real-world tests at the training field of a fire department. The results of these validation tests closely followed Karo's performance in RRL, where about 77% of all 13 tests successfully met the mission objectives. However, we observed a few shortcomings in our system that captured useful information for future improvements: (1) limited resiliency to carry various devices that could be highly pragmatic in a field operation, (2) insufficient robustness in climbing stairs with harsh debris, and (3) inadequate non-line-of-sight communication, which diminished the overall performance.

Future work will aim to address the deployment and maintenance necessities and apply them thoroughly into the design process. The improvement of the manipulation system will continue to increase payload and object handling. Another direction will be focused on implementing the object recognition feature to classify the perceived information for the operator. Future studies will also be focused on enhancing autonomous exploration by using the IMU orientation data to develop an advanced scheme to control the flippers. To that end, the control scheme will be trained with a training data set obtained from an expert human–robot operator by utilizing supervised machine learning techniques, which effectively facilitates the robot's navigation in simple repeatable missions [52, 53].

### Abbreviations
USAR: Urban Search and Rescue; RRL: Rescue Robot League; DOF: Degrees of Freedom; DHS S&T: U.S. Department of Homeland Security, Science and Technology Directorate; NIST: National Institute of Standard and Technology; HRI: Human–Robot Interface; AMRL: Advanced Mobile Robotics Lab; ASTM: American Society for Testing and Materials; COM: Center of Mass; FBD: Free Body Diagram; OCU: Operator Control Unit; SLAM: Simultaneous localization and mapping; GUI: Graphical User Interface; CAN: Controller Area Network; ROS: Robot Operating System.

### Acknowledgement
The authors would like to acknowledge the work of the team members at AMRL who offered their enthusiasm and collaborative support to the Karo project [54].

### Authors' contributions
SH, MD, BP, AH, MHS, and AHMH conducted research and experiments as active members of AMRL. SH and MD also analyzed the data, developed the key studies, and wrote the manuscript. FN advised the research and experiments of this project. All authors read and approved the final manuscript.

### Funding
This research was funded by Qazvin Azad University via internal grants to assist the project at the Advanced Mobile Robotics Lab (AMRL) [54].

### Availability of data and materials
The data and source code used to support the findings of this study are available from the corresponding author upon request.

### Competing interests
The authors declare that they have no competing interests.

### Author details
[1] The Authors Were With Advanced Mobile Robotics Lab, Qazvin Azad University, Qazvin, Iran. [2] University of Tehran, College of Engineering, School of Mechanical Engineering, Tehran, Iran.

## Publisher's Note